\documentclass[sort&compress,5p,twocolumn]{elsarticle} 
\usepackage{lineno,hyperref}
\usepackage[dvipsnames]{xcolor}
\usepackage{picins}
\usepackage{adjustbox}
\usepackage[caption=false,font=scriptsize]{subfig}

\usepackage{relsize}

\journal{Neurocomputing}

\usepackage{amssymb}

\usepackage{graphicx}
\usepackage{mathptmx}      

\usepackage{dsfont}
\usepackage{url}

\usepackage{amssymb}

\usepackage[cmex10]{amsmath}

\usepackage{amssymb}
\usepackage{algorithm}
\usepackage{algorithmic}

\usepackage{pgfplots}
\pgfplotsset{compat=newest}

\usepackage{subfloat}

\usepackage[latin1]{inputenc} 
\usepackage{graphicx}
\usepackage{amsmath,epsfig}
\usepackage{amssymb,cases}
\usepackage{colortbl}
\usepackage[T1]{fontenc}
\usepackage{etex}
\usepackage[all,knot,arc,frame,matrix,arrow,graph,curve,color]{xy}
\usepackage{amssymb,amsthm}
\usepackage{leftidx}
\usepackage{graphics}

\usepackage{booktabs}
\usepackage{bigstrut}
\usepackage{multirow}
\usepackage{tkz-graph}
\usepackage{xcolor}
\usepackage{tikz}
\usepackage{verbatim}

\usetikzlibrary{shapes}
\usetikzlibrary{positioning,arrows}
\usetikzlibrary{patterns}

\usepackage{pgfplots}

\makeatother

\pgfkeys{
        /tr/rowfilter/.style 2 args={
                /pgfplots/x filter/.append code={
                        \edef\arga{\thisrow{#1}}
                        \edef\argb{#2}
                        \ifx\arga\argb
                        \else
                                
                        \fi
                }
        }
}

\pgfplotsset{
    discard if/.style 2 args={
        x filter/.code={
            \edef\tempa{\thisrow{#1}}
            \edef\tempb{#2}
            \ifx\tempa\tempb
                
            \else
            \fi
        }
    }
}

\pgfplotsset{
    discard if not/.style 2 args={
        x filter/.append code={
            \edef\tempa{\thisrow{#1}}
            \edef\tempb{#2}
            \ifx\tempa\tempb
            \else
                
            \fi
        }
    }
}

\makeatletter
\pgfplotstableset{
    discard if not/.style 2 args={
        row predicate/.append code={
            \def\pgfplotstable@loc@TMPd{\pgfplotstablegetelem{##1}{#1}\of}
            \expandafter\pgfplotstable@loc@TMPd\pgfplotstablename
            \edef\tempa{\pgfplotsretval}
            \edef\tempb{#2}
            \ifx\tempa\tempb
            \else
                \pgfplotstableuserowfalse
            \fi
        }
    }
}
\makeatother

\makeatletter
\pgfplotstableset{
    select if/.style 2 args={
        row predicate/.append code={
            \def\pgfplotstable@loc@TMPd{\pgfplotstablegetelem{##1}{#1}\of}
            \expandafter\pgfplotstable@loc@TMPd\pgfplotstablename
            \edef\tempa{\pgfplotsretval}
            \edef\tempb{#2}
            \ifx\tempa\tempb
						\pgfplotstableuserowtrue
						\else
            \fi
        }
    }
}
\makeatother

\newcounter{definicao}
\renewcommand \thedefinicao
     {\ifnum \c@section>\z@ \fi \@arabic\c@definicao}
\def\fps@definicao{tbp}
\def\ftype@definicao{3}
\def\ext@definicao{lod}
\def\fnum@definicao{\definicaoname~\thesection.\thedefinicao}

\newcounter{contProgram}

\newenvironment{definicao*}
               {\@dblfloat{definicao}}
               {\end@dblfloat}

\newcommand\definicaoname{{\bf Definition}}

\newcommand\Comment[1]{{}}

\newcounter{corolario}[section]
\renewcommand \thecorolario
     {\ifnum \c@chapter>\z@ \fi \@arabic\c@corolario}
\def\fps@corolario{tbp}
\def\ftype@corolario{3}
\def\ext@corolario{lod}
\def\fnum@corolario{\corolarioname~\thesection.\thecorolario}

\newcounter{contCorolario}[section]

\newenvironment{corolario*}
               {\@dblfloat{corolario}}
               {\end@dblfloat}

\newcommand\corolarioname{{\bf Corolary}}

\newcommand\First[1]{{\mbox{First}(#1)}}

\tikzstyle{background2}=[rectangle,fill=black!10!white,inner sep=0.2cm,rounded corners=1mm]
\tikzstyle{vertexNULL}=[rectangle,minimum size=20pt,inner sep=0pt]
\tikzstyle{vertex}=[circle,fill=black!25,minimum size=20pt,inner sep=0pt]

\tikzstyle{vertexG}=[circle,fill=black!40,text=white,minimum size=15pt,inner sep=0pt]
\tikzstyle{edgeG} = [draw,thick,-]
\tikzstyle{edgeMG} = [draw,dotted,thick,-]
\tikzstyle{edgeEMG} = [draw,dashed,thick,-]

\tikzstyle{order}=[rectangle,fill=blue!5,minimum size=20pt,inner sep=0pt]
\tikzstyle{vertexNAME}=[rectangle,minimum size=20pt,inner sep=0pt]

\tikzstyle{vertexCT}=[rectangle,fill=green!05,minimum size=20pt,inner sep=0pt]
\tikzstyle{vertexKCT}=[circle,fill=red!45,minimum size=15pt, inner sep=0pt]
\tikzstyle{selected vertex} = [vertex, fill=red!24]
\tikzstyle{partition 1} = [vertex, fill=red!24]
\tikzstyle{partition 2} = [vertex, fill=blue!24]
\tikzstyle{partition 3} = [vertex, fill=green!24]
\tikzstyle{partition 4} = [vertex, fill=gray!24]
\tikzstyle{partition 5} = [vertex, fill=red!70]
\tikzstyle{partition 6} = [vertex, fill=yellow!50]
\tikzstyle{partition 7} = [vertex, fill=yellow!20]

\tikzstyle{edgeCT} = [draw,thick,-,red!50]
\tikzstyle{edge} = [draw,line width=7pt,-,blue!50]
\tikzstyle{edge initial} = [draw,thick,-]
\tikzstyle{edge initial CT} = [draw,thick,->]
\tikzstyle{edge initial KCT} = [draw,thick,->,red!20]

\tikzstyle{weight} = [font=\small]
\tikzstyle{selected edge} = [draw,line width=5pt,-,red!50]
\tikzstyle{ignored edge} = [draw,line width=5pt,-,black!20]

\tikzstyle{abstractSJ}=[rectangle, draw=black, rounded corners, fill=blue!40, drop shadow,
        text centered, anchor=north, text=white, text width=0.5cm,minimum size=20pt]
\tikzstyle{commentSJ}=[anchor=center,text=white]
\tikzstyle{myarrowSJ}=[->, >=open triangle 90, thick]
\tikzstyle{lineSJ}=[-, thick]

\tikzstyle{vertexNULL}=[rectangle,minimum size=20pt,inner sep=0pt]
\tikzstyle{vertex}=[circle,fill=black!25,draw=black,minimum size=20pt,inner sep=0pt]
\tikzstyle{order}=[rectangle,fill=blue!5,minimum size=20pt,inner sep=0pt]

\tikzstyle{vertexRegionA}=[circle,fill=black!10,draw=black,minimum size=20pt,inner sep=0pt]
\tikzstyle{vertexRegionB}=[circle,fill=white!100,draw=black,minimum size=20pt,inner sep=0pt]
\tikzstyle{vertexRegionC}=[circle,fill=black!50,text=white,draw=black,minimum size=20pt,inner sep=0pt]
\tikzstyle{vertexRegionD}=[circle,fill=black!90,text=white,draw=black,minimum size=20pt,inner sep=0pt]

\tikzstyle{vertexCT}=[rectangle,fill=green!05,minimum size=20pt,inner sep=0pt]
\tikzstyle{vertexKCT}=[circle,fill=red!45,minimum size=15pt, inner sep=0pt]
\tikzstyle{selected vertex} = [vertex, fill=red!24]
\tikzstyle{partition 1} = [vertex, fill=red!24]
\tikzstyle{partition 2} = [vertex, fill=blue!24]
\tikzstyle{partition 3} = [vertex, fill=green!24]
\tikzstyle{partition 4} = [vertex, fill=gray!24]
\tikzstyle{partition 5} = [vertex, fill=red!70]
\tikzstyle{partition 6} = [vertex, fill=yellow!50]
\tikzstyle{partition 7} = [vertex, fill=yellow!20]

\tikzstyle{edgeCT} = [draw,thick,-,red!50]
\tikzstyle{edge} = [draw,line width=7pt,-,blue!50]
\tikzstyle{edge initial} = [draw,thick,-]
\tikzstyle{edge initial CT} = [draw,thick,->]
\tikzstyle{edge initial KCT} = [draw,thick,->,red!20]

\tikzstyle{weight} = [font=\small]
\tikzstyle{selected edge} = [draw,line width=5pt,-,red!50]
\tikzstyle{ignored edge} = [draw,line width=5pt,-,black!20]

\usepackage{pgfplots}
\usepackage{pgfplotstable}
\usepackage{pgfplots, pgfplotstable}
\usepackage{siunitx}
\usepgfplotslibrary{ternary}

\pgfkeys{
        /tr/rowfilter/.style 2 args={
                /pgfplots/x filter/.append code={
                        \edef\arga{\thisrow{#1}}
                        \edef\argb{#2}
                        \ifx\arga\argb
                        \else
                                
                        \fi
                }
        }
}

\pgfplotsset{
    discard if/.style 2 args={
        x filter/.code={
            \edef\tempa{\thisrow{#1}}
            \edef\tempb{#2}
            \ifx\tempa\tempb
                
            \else
            \fi
        }
    }
}

\usetikzlibrary{external}
\newcommand{\red}[1]{\textcolor{black}{#1}}

\title{A mid-level video representation based on binary descriptors: \\ A case study for pornography detection}

\begin{document}
	\begin{frontmatter}
		
	\author[ufmg-npdi,ufmg-ssig]{Carlos~Caetano\corref{correspondingauthor}}
	\ead{carlos.caetano@dcc.ufmg.br}
    \cortext[correspondingauthor]{Corresponding author.}
	
	\author[unicamp]{Sandra~Avila}
	\ead{sandra@dca.fee.unicamp.br}
	
	\author[ufmg-ssig]{William~Robson~Schwartz}
	\ead{william@dcc.ufmg.br}
	
	\author[pucmg]{Silvio~Jamil~F.~Guimar\~aes}
	\ead{sjamil@pucminas.br}
	
	\author[ufmg-npdi]{Arnaldo~de~A.~Ara\'ujo}
	\ead{arnaldo@dcc.ufmg.br}

	\address[ufmg-npdi]{Universidade Federal de Minas Gerais, NPDI --- DCC/UFMG, Minas Gerais, Brazil}
	\address[ufmg-ssig]{Universidade Federal de Minas Gerais, SSIG Group --- DCC/UFMG, Minas Gerais, Brazil}
	\address[unicamp]{University of Campinas, RECOD Lab --- DCA/FEEC/UNICAMP, Campinas, Brazil}
	\address[pucmg]{Pontifical Catholic University of Minas Gerais, VIPLAB --- ICEI/PUC Minas, Minas Gerais, Brazil}
		
    \begin{abstract}
    With the growing amount of inappropriate content on the Internet, such as pornography, arises the need to detect and filter such material. The reason for this is given by the fact that such content is often prohibited in certain environments (e.g., schools and workplaces) or for certain publics (e.g., children). In recent years, many works have been mainly focused on detecting pornographic images and videos based on visual content, particularly on the detection of skin color. Although these approaches provide good results, they generally have the disadvantage of a high false positive rate since not all images with large areas of skin exposure are necessarily pornographic images, such as people wearing swimsuits or images related to sports. Local feature based approaches with Bag-of-Words models (BoW) have been successfully applied to visual recognition tasks in the context of pornography detection. Even though existing methods provide promising results, they use local feature descriptors that require a high computational processing time yielding high-dimensional vectors. In this work, we propose an approach for pornography detection based on local binary feature extraction and BossaNova image representation, a BoW model extension that preserves more richly the visual information. Moreover, we propose two approaches for video description based on the combination of mid-level representations namely BossaNova Video Descriptor (BNVD) and BoW Video Descriptor (BoW-VD). The proposed techniques are promising, achieving an accuracy of 92.40\%, thus reducing the classification error by 16\% over the current state-of-the-art local features approach on the Pornography dataset.
    \end{abstract}
		
		
	\end{frontmatter}
	
	
	\section{Introduction}\label{introduction}

Due to the fast growth of images and videos publicly available on the Internet, the need for recognition of their contents arises. Besides the obvious need for methods related to image and video searches, it is also important to perform recognition or classification of contents that may be considered undesirable or offensive to allow the development of methods for filtering~them.


The largest group of images and videos available on the Internet that people may find offensive is related to pornographic materials. A report by the ExtremeTech\footnote{\url{http://www.extremetech.com/computing/123929-just-how-big-are-porn-sites}} technology site suggests that 30\% of all Internet traffic is associated with pornography. They arrived at this number by estimating the traffic that a popular pornographic website generates every day and multiplied it by several other pornographic websites of similar size found on the Internet. According to their report, the largest website provider of this type of content receives three times more pageviews than major news websites (about 4.4 billion pageviews per month) and the average time spent on this site can be five times higher than in news sites.

According to Short et al.~\citep{Short:2012}, pornography can be considered as any sexually explicit material with the aim of sexual arousal or fantasy. However, this definition leads to many challenges when trying to detect pornographic content, such as the bounds of ``explicit'' for something to be considered as pornographic material. Some works in the literature deal with this issue by dividing the material into several classes~\citep{Deselaers:2008}, complicating even more the classification task. On the other hand, there are works that choose to deal with it by using a conceptually simple evaluation considering only two classes (pornographic and non-pornographic) \citep{Valle:2011, Avila:2013a}, the focus of this work.

Detecting and filtering pornographic visual content from the Internet is a concern in many environments (e.g., schools, workplaces). According to Lopes et al.~\citep{Lopes:2009a}, linked text tags to pictures and videos are clearly not sufficient, since inappropriate content can be maliciously attached to seemingly innocent texts. A typical situation would be, for example, the employment of search keywords commonly used by children attached to websites with pornographic content. In addition, adults may also not wish to be exposed to such contents, for \red{instance}, from results received from search engines available on the web.

In recent years, several works in literature have been mainly focused on detecting pornographic images and videos based on visual content rather than textual information \citep{Deselaers:2008, Valle:2011, Forsyth:1996, Forsyth:1997, Forsyth:1999, Jones:2002, Zheng:2004, Rowley:2006, Hu:2007, Lopes:2009b, Ulges:2011, Steel:2012, Avila:2013a, Yu:2014, Caetano:2013, Zhuo:2015}. Most of these works are based on skin color detection approaches since a large fraction of pixels that have colors related to the human skin \citep{Ries:2012}. Nevertheless, a shortcoming of these approaches is related to the high rate of false positives, since not all images with large areas of skin exposure are necessarily pornographic (pictures of people wearing swimsuits, or sports-related images). Furthermore, another issue to be considered is that grayscale pictures \red{cannot} be classified using color related~features.

The pornography detection task can be interpreted as a visual recognition task in the context of object \red{recognition}~\citep{Lopes:2009a}. Approaches based on local features in conjunction with Bag-of-Words models (BoW) have been successfully applied to visual classification tasks \citep{Agarwal:2004, Yang:2007}. In such approaches, images are represented as histograms constructed from a set of visual features. No explicit model of the object is needed and the variability of examples (related to \red{rotation}, shape scale or illumination) is treated by a training set that includes such variability. In view of that, approaches based on BoW models are suitable to the task of pornography detection.


Despite the existing methods based on BoW models produce promising results in the pornography detection context, these also make use of local feature descriptors that require a high computational processing time and generate high-dimensional real-valued vectors. For example, Avila et al.~\citep{Avila:2013a} made use of HueSIFT feature descriptor \citep{HueSIFT}, a variant of SIFT descriptor \citep{Lowe:2004} that includes color information, taking an average time of 2.5 seconds to densely extract the local features of an image generating a feature vector consisting of 165 floating point values. In fact, this is still not fast enough for \red{real-time} applications, that require a \red{short} response time. Moreover, the comparison between two extracted features would spend more computational time due to the high dimensionality. On the other hand, to satisfy the requirements of web pornographic image recognition both on precision and speed, Zhuo et al.~\citep{Zhuo:2015} proposed a pornographic image recognition method based on the binary descriptor Oriented FAST and Rotated BRIEF (ORB), which is a low-complexity alternative. However, their work focused only on static images.

In this paper, we formalize a video descriptor approach to the visual recognition problem in the context of pornography detection in videos. 
The method is based on \red{both} a low-complexity alternative for feature extraction using binary descriptors and \red{a} combination of mid-level representations. We apply it to the classical BoW model generating the BoW Video Descriptor (BoW-VD). We also apply it to the BossaNova, a BoW model extension that preserves the visual information in a richer way, which generates the BossaNova Video Descriptor (BNVD). Our proposal has as advantage the fact that it does not depend on any skin detector or shape models to classify pornography; besides, according to the experimental results, it outperforms the state-of-the-art results on the Pornography dataset \citep{Avila:2013a}. \emph{To the best of our knowledge, ours is the best result reported to date on the Pornography dataset employing local feature descriptors.}

The use of binary descriptors and the mid-level representation for videos were first introduced in our previous works~\cite{Caetano:2013} and~\cite{Caetano:2014:EUSIPCO}. This paper presents several new aspects in comparison with the previous ones. Those aspects are highlighted in the following:

\begin{itemize}
\item \red{Formalization} of BossaNova Video Descriptor (BNVD). In this work, we present a \red{new} formulation which generalizes the BNVD allowing the use of different aggregation functions.

\item Proposal of BoW Video Descriptor (BoW-VD). In this work, we also propose a video descriptor by using aggregation functions for combining the traditional BoW mid-level \red{representations}.

\item Improvement of the experimental results. In this work, we study the behavior of binary descriptors and the mid-level representation by using several parameter settings, including the use of global pooling for creating a video descriptor. 
Moreover, we show the computational times for \red{generating} our video descriptor.
\end{itemize}

The remainder of this paper is organized as follows. We start by explaining the classical non-binary local feature descriptors, the most recent binary feature descriptors and the BossaNova mid-level representation (Section~\ref{theoreticalbackground}). Next, we survey the recent works on pornography detection (Section~\ref{relatedworks}). We then introduce the complete formalism of our video descriptor (Section~\ref{proposal}). Afterwards, we analyze our experiments regarding the proposed video descriptor and we perform a comparison with state-of-the-art approaches (Section~\ref{experiments}). Finally, we present our concluding remarks (Section~\ref{conclusion}). 
\section{Theoretical Background}\label{theoreticalbackground}

The most common approach for visual recognition \red{task} consists of three distinct steps~\citep{Chatfield:2011}: (i) extraction of local features; (ii) encoding of the local features in an intermediate representation (mid-level); and (iii) classification of the mid-level representation, usually based on machine learning techniques. Typically, the extracted local features tend to be invariant to some transformations caused by camera changes such as rotation, scale and illumination. To address these properties, local features \red{such as SIFT \citep{Lowe:2004} or SURF \citep{SURF} descriptors are usually extracted}. Regarding the mid-level representation, BoW models \citep{Sivic:2003} are the most common approaches used to encode the extracted local features. \red{Moreover, to improve the efficiency of retrieval and classification without sacrificing accuracy, there are several methods \cite{Datar:2004, Gong:2011, Shen:2014, Shen:2015, Santos:2015} that can be applied to the local features or mid-level representations in order to convert them into compact similarity-preserving binary codes.} Finally, the purpose of the classification is to learn a function able to assign discrete labels to images/videos. To that end, most of the visual recognition works make use of machine learning techniques, such as Support Vector Machines (SVM).

\subsection{Local Feature Descriptors}

A local feature descriptor can be considered as a function applied to a region of the image to perform its description. The simplest way to describe a region is to represent all the pixels in this region in a single vector. However, depending on the \red{information} to be described, this would result in a high-dimensional vector leading also to a high computational complexity for a future recognition of this region~\citep{GLOH}. 

In this section, we review local descriptors, which can be classified in two distinct ways \citep{Canclini:2013}: (i) non-binary descriptors and (ii) binary descriptors. It is important to say that new approaches for local descriptors have been proposed in the literature, so the following list is not an exhaustive list. 
However, it can be considered as a representative group of the most relevant descriptors for our context.

\subsubsection{Non-Binary Descriptors}

\subsubsection*{SIFT -- Scale Invariant Feature Transform}

One of the most important descriptors used in the literature is the SIFT~\citep{Lowe:2004}. This descriptor performs a scale-space analysis leading to a great performance according to the scale invariance \citep{Morel:2011}. Although the author has developed the SIFT descriptor to be used on object recognition tasks, it has become the most widely used descriptor in several other applications. This is due to its high discriminative power and stability.

\red{To describe each patch}, an orientation $\alpha$ is assigned selecting the angle that represents the histogram of local gradients (calculated for each pixel around the keypoint). Then, the region of points around the keypoint, oriented by $\alpha$, is divided into \red{subregions} composed by a grid of size $G \times G$. \red{Next,} a histogram of orientation consisting of $B$ bins is created from the samples of each \red{subregion}. The descriptor is then obtained from the concatenation of the histograms of these \red{subregions}, composed of $G \times G \times B$ values. The default values for $G$ and $B$ are usually 4 and 8, respectively, resulting in a vector of 128 length. Finally, the descriptor is normalized turning it robust to illumination variations.

\subsubsection*{SURF -- Speeded-Up Robust Features}

To overcome the problem of high computational processing time of SIFT, Bay et al.~\citep{SURF} proposed a faster descriptor, called SURF. \red{It} can be seen as an approximation of SIFT and \red{it} has the same idea of using histograms based on local gradients. SURF is based on integral images to approximate convolutions, which provides a considerable improvement in efficiency (as compared to SIFT). Despite the approximations on the descriptor creation, there is no considerable loss in the rotation and scale invariance.

Similarly to SIFT, SURF assigns an orientation to each patch described: a circular region around the keypoint is described according to the distribution of responses received by a Haar-wavelet filter. The size of the wavelet region and the sampling parameter are dependent of a scale $\sigma$ in which the keypoint is detected. Weighted with a Gaussian function around the keypoint, the filter responses are represented by vectors in a two-dimensional space and then summed up. The dominant orientation determines the orientation of the keypoint. Then, the patch is divided into a grid consisting of $4 \times 4$ \red{subregions}. For each \red{subregion}, a feature vector composed of four dimensions is calculated using a Haar wavelet-filter and then a sum vector of orientations is calculated in each cell. Finally, the concatenation of the feature vectors of each \red{subregion} produces the SURF descriptor ($4 \times 4 \times 4 = 64$ dimensions).

\subsubsection{Binary Descriptors}

\subsubsection*{BRIEF -- Binary Robust Independent Elementary Features}

\red{The BRIEF descriptor \citep{BRIEF} is one the simplest of the binary descriptors and also the first published.} By itself, BRIEF is neither scale nor rotation invariant and does not have an elaborate sampling pattern. Nevertheless, its performance is similar to a more complex local descriptor, the SURF, when compared to its robustness to illumination, blur, and perspective distortion.

The BRIEF descriptor is represented by a binary string in which each bit represents a simple comparison between two elements inside a patch. 
The bit is set to `1' if the first point is more intense than the other one, otherwise it is set to `0'. 
The most common strategy for choosing these points is based on a randomly way according to a Gaussian distribution with respect to the keypoint of the patch. The number of selected points leads to the descriptor size (e.g., 128, 256 and 512 \red{bits}).

\subsubsection*{ORB -- Oriented FAST and Rotated BRIEF}

The ORB descriptor \citep{ORB} is similar to BRIEF. However, it is robust to noise and invariant to rotation. The invariance to rotation is achieved by estimating the patch rotation using the intensity centroid. Patch moments are used to compute the intensity centroid, which outperform gradient-based approaches.

The sampling pattern is steered estimating the orientation and usual binary tests are used for computing the descriptor. Furthermore, for selecting a couple of points, a $k$-nearest neighborhood strategy based on error-prone is performed. The random sampling has been replaced by a sampling scheme that uses machine learning for decorrelating BRIEF features under rotational invariance. Unlike BRIEF, the ORB descriptor size is fixed in 256 bits.

\subsubsection*{BRISK -- Binary Robust Invariant Scalable Keypoints}


The approach used by BRISK~\citep{BRISK} is very similar to BRIEF once the descriptor is computed based on simple binary tests comparing pixel intensities. However, BRISK presents three main differences: (i) it takes into account the rotation of the point to be described; (ii) it makes use of a scale-space theory to maximum adapt the sampling pattern in scale space; and (iii) it uses a special pattern for the binary tests, instead of a random probability distribution. In this way, BRISK becomes invariant to rotation and scale.

\begin{figure}[tb]
	\centering
	\subfloat[BRISK~\citep{BRISK}]{\includegraphics[width=0.45\linewidth]{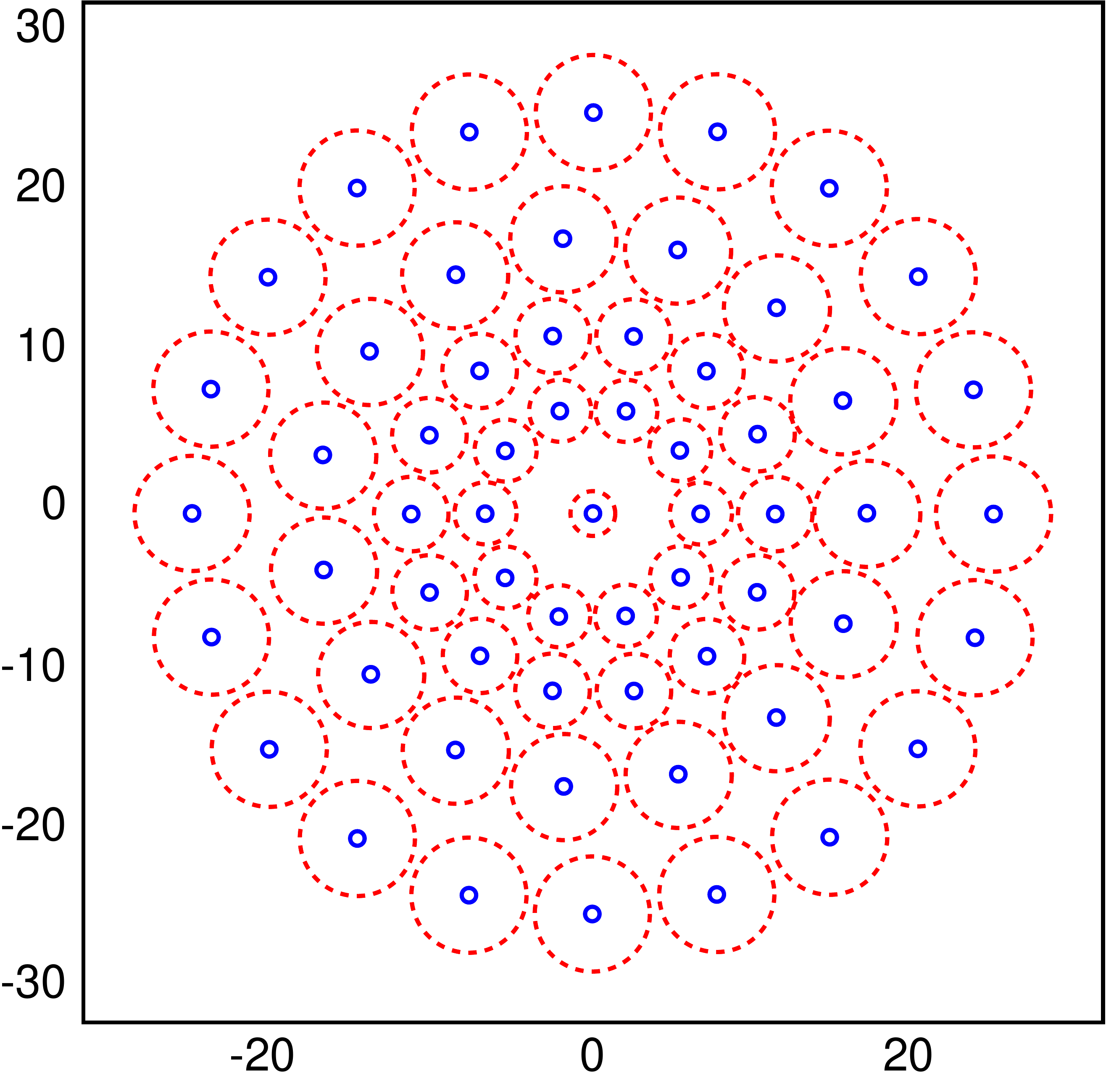}}\label{fig:sampling:patterns:a} \hspace{0.2cm}
	\subfloat[FREAK \citep{FREAK}]{\includegraphics[width=0.45\linewidth]{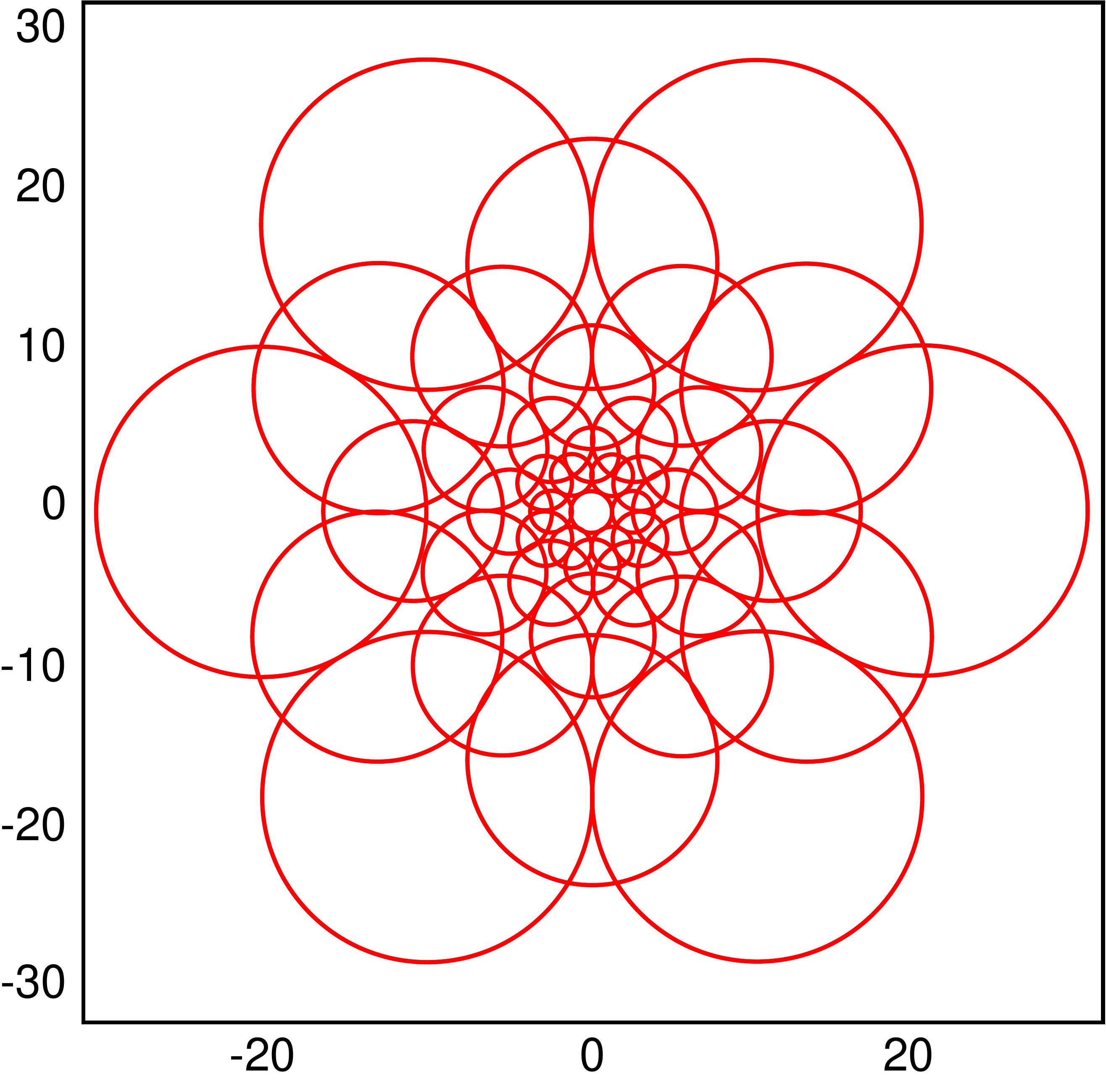}}\label{fig:sampling:patterns:b} \\
	\caption{Sampling patterns of two local binary descriptor. (a) Illustrates the sampling pattern of BRISK descriptor which is based on 60 points: the small blue circles denote the sampling locations; the bigger red dashed circles are drawn at a radius, which corresponds to the standard deviation of the Gaussian kernel used to smooth the intensity values at the sampling points. (b) Illustrates the sampling pattern of FREAK descriptor in which each circle represents a receptive field where the image is smoothed with its corresponding Gaussian kernel. }
	\label{fig:sampling:patterns}
\end{figure}

As illustrated in Figure~\ref{fig:sampling:patterns}(a), the BRISK descriptor uses a pattern of points $p_{i}$ equally distributed in concentric circles around the keypoint to be described. To compare points, the authors defined two distinct sets of pairs of points, long distance pairs and short pairs. Long distance pairs are composed of $(i,j)$ where $\left\|p_{i} - p_{j}\right\| > \delta_{min}$ and \red{they} are used to estimate the orientation of the keypoint using a gradient mean. Then, a Gaussian smoothing is applied to the concentric rings used for the sampling pattern and 512 short pairs (whose distance is less than a threshold $\delta_{max}$) are used to build the descriptor with simple comparisons between the points.

\subsubsection*{FREAK -- Fast Retina Keypoint}




To extend the BRISK descriptor, Alahi et al.~\citep{FREAK} proposed a descriptor called FREAK. Similarly to BRISK, the FREAK descriptor has its sampling pattern based on Gaussians but its sampling point distribution is biologically inspired by the retina pattern of the human eye.

FREAK has two main important differences compared to BRISK. First, it consists of an allocation of concentric distributions with an exponential growth over the distance of the keypoint. The second difference is based on the fact that the sampling pattern overlaps on different concentric circles, as shown in Figure~\ref{fig:sampling:patterns}(b). The overlap between sampling regions adds some redundancy that increases the discriminative power of the descriptor. The authors mention that this redundancy is also present in the receptive fields of the human retina.

The sampling pattern used in the original implementation of FREAK descriptor consists of 43~``receptive fields'', leading to 903~possible comparison tests. Thus, for the construction of the final descriptor, FREAK uses a similar method to the ORB descriptor using a greedy approach to select the less correlated comparison tests making it more discriminative. To achieve a maximum performance, 512~binary tests are used.

\subsubsection*{BinBoost}

Trzcinski et al.~\citep{BinBoost} proposed a new framework with the aim of creating a binary descriptor extremely compact and highly discriminative. \red{BinBoost descriptor} is robust to changes in lighting and viewpoint.

Unlike the aforementioned binary descriptors which compute the feature vector based on simple binary tests comparing the intensity of pixels, each bit generated by BinBoost is computed by using a binary hash function, \red{the same way as the AdaBoost classifier does \citep{AdaBoost}}. This function is based on weak learners that take into account the orientation intensity of gradients on the patch to be described. The hash function is optimized iteratively, i.e., at each iteration, incorrect samples are assigned to a greater weight while the weight of the correct samples is decreased. In this way, the next bits to be calculated will tend to correct the error of their predecessors.

\subsection{Mid-level Representation: BossaNova}

We now overview the BossaNova mid-level image representation, which introduces a density-based pooling strategy by computing the histogram of distances between the local descriptors and the codewords. More details can be found in~\citep{Avila:2013a,Avila:2011}.

Let ${\mathcal X}$~be an unordered set of binary descriptors extracted from an image. ${\mathcal X} = \left\{\mathbf{x}_j \right\}$, $j \in \left[1,N\right]$, where ${\mathbf{x}_j\in \{0,1\}^D}$ is a binary descriptor vector and $N$ is the number of binary descriptors in the image.
Let ${\mathcal C}$~be a visual codebook\footnote{The codebook is usually built by clustering a set of local descriptors. It can be defined by the set of codewords (or visual words) corresponding to the centroids of clusters.} obtained by the $k$-medians algorithm. ${\mathcal C} = \left\{ \mathbf{c}_m\right\}$, $m \in \left[1,M\right]$, where $\mathbf{c}_m \in \{0,1\}^D$ is a codeword and $M$ is the number of visual codewords. $\mathbf{z}$ is the final vectorial BossaNova representation of the image used for classification.

To keep more information than the BoW during the pooling step, the BossaNova pooling function,~$g$, estimates the probability density function of $\alpha_{m}$: $g(\alpha_{m}) = \operatorname{pdf}( \mathbf{\alpha_{m}})$, by computing the following histogram of distances $z_{m,b}$:
\vspace{-0.1cm}
\begin{eqnarray}\label{eq:bossa}
g : \mathds{R}^N &\longrightarrow& \mathds{R}^B,      \nonumber \\
    \mathbf{\alpha_{m}} &\longrightarrow& g(\alpha_{m}) = z_m,  \nonumber \vspace{0.2cm}\\
    z_{m,b} &=&\operatorname{card} \Big(\mathbf{x}_j~|~\alpha_{m,j}  \in  \Big[\frac{b}{B}; \dfrac{b+1}{B}\Big]\Big), \nonumber \\
    & & \frac{b}{B} \geq \alpha_m^{min} ~~\text{and}~~  \dfrac{b+1}{B} \leq \alpha_m^{max}, 
\end{eqnarray}

\noindent where $B$ denotes the number of bins of each histogram~$z_m$, $\alpha_{m,j}$ represents a dissimilarity (i.e., a distance) between $\mathbf{c}_m$ and $\mathbf{x}_j$, and $[\alpha_m^{min};\alpha_m^{max}]$ limits the range of distances for the descriptors considered in the histogram computation.

In addition to the pooling strategy, Avila et al.~\citep{Avila:2013a} also proposed a localized soft-assignment coding that considers only the $k$-nearest codewords for coding a local descriptor.
After computing a local histogram~$z_m$ for all the~$c_m$ centers, the BossaNova image representation $\mathbf{z}$ \citep{Avila:2013a} is given by:
\vspace{-0.1cm}
\begin{equation}
\label{eq:bossanova}
 \mathbf{z} = \left[ \left[z_{m,b}\right], st_m\right]^\text{T},~~(m,b) \in \{1,...,M\} \times \{1,...,B\}, 
\end{equation}

\noindent where $\mathbf{z}$ is a vector of size $M \times (B+1)$, $s$ is a nonnegative constant and $t_m$ is a scalar value for each codeword, counting the number of binary descriptors $\mathbf{x}_j$ close to that codeword. 

The idea of enriching BoW representations with extra knowledge from the set of local descriptors has been explored on several approaches~\cite{Sanchez:2013,Zhou:2010}. However, those works opt by parametric models that lead to very high-dimensional image representations. By using a simple histogram of distances to capture the relevant information, BossaNova remains flexible and keeps the representation compact. For these reasons, we decided to employ it in this work for mid-level features.

In short, the BossaNova vector is defined by three parameters: (i) the number of codewords $M$; (ii) the number of bins $B$ in each histogram; and  (iii) the range of distances $[\alpha_m^{min}, \alpha_m^{max}]$.
As in \citep{Avila:2013a}, we set up the bounds as $\alpha_m^{min}= \lambda_{min} \cdot \sigma_m $ and $\alpha_m^{max}= \lambda_{max} \cdot \sigma_m$, where $\sigma_m$ is the standard deviation of each cluster $\mathbf{c}_m$ obtained by $k$-medians clustering algorithm.


	\section{Related Works}\label{relatedworks}

According to Ries and Lienhart~\cite{Ries:2012}, works involving pornography detection can be divided into three main groups: (i) approaches based on skin color that exploit the assumption that pornographic images/videos generally have large areas of skin color; (ii) approaches based on shape information; and (iii) approaches based on local features in conjunction with Bag-of-Words (BoW) models. 

In this \red{section}, we survey the literature on the pornography detection. In \red{Section}~\ref{subsec:skincolorshape}, we cover works based on skin color detection as well as works based on shape information, since all shape based approaches presented in this paper also employ a step of finding pixels that have skin related colors. \red{Section}~\ref{subsec:localfeatures} presents the works that make use of local features and BoW models.

\subsection{Skin Color and Shape Information Based Approaches} \label{subsec:skincolorshape}

Most of the pornography detection works are based on skin color detection or shape information. This is partly because the most obvious property in pornographic images is the large fraction of pixels presenting skin related colors and that most of the pornographic images share some characteristic shapes~\cite{Ries:2012}.

The approach proposed by Forsyth and Fleck~\cite{Forsyth:1999} begins by finding areas with skin color in the image. First, they \red{transformed} each pixel value into intensity value and two tone values. After that, decision rules are applied to find regions with skin color. Upon skin detection, a corner detector and a Hough transform are applied to find candidates for human limbs. These candidates are iteratively combined according to a set of constraints to model the geometry of the human body. Finally, if it is possible to assemble the limbs in a geometrically reasonable way, the image is classified as pornographic.

Jones and Rehg~\cite{Jones:2002} constructed a 3D histogram of 256 bins for each color channel. From these histograms, five different features are extracted, such as the percentage of pixels related to the skin or the number of connected areas of skin. Finally, a decision tree is trained based on these characteristics. However, the authors presented results suggesting that histogram with less bins are sufficient, since the latter can cause overfitting.

Inspired by the color histogram used by Jones and Rehg~\cite{Jones:2002}, Rowley et al.~\cite{Rowley:2006} generated a skin color based map to determine connected components. Then, skin based and connected component features are extracted from the map, such as mean and standard deviations. The authors also employed other color characteristics (e.g., the edge pixels within the skin regions). Finally, these characteristics are used as input to a \red{SVM}~classifier.

The approach presented by Lee et al.~\cite{Lee:2007} employed a learning scheme based on the skin color distribution of the image, using a neural network to learn and classify whether the input image contains skin exposure. Furthermore, a feature is used to detect textures with roughness to reject non-skin objects. Then, three types of features related to the targeted form (area size, aspect ratio and location) are extracted and sent to an AdaBoost classifier. Finally, a face detection algorithm is applied to filter out false candidates related to face photos. 
Using the latter technique, Lee et al.~\cite{Lee:2013} separated the image colors in skin and non-skin groups. Next, a texture analysis is applied to verify if the likelihood of the area is composed by skin and a face detection algorithm is applied to eliminate face photos. In addition, the authors verified the presence of ``holes'' in the binary images to detect photos related to swimsuits. For the remaining images, features regarding the position of the skin region and morphological characteristics are extracted to train a SVM~classifier.

A method to estimate skin regions was proposed by Yu and Han~\cite{Yu:2014} using simple operations in the HSV color space plus an additional post-processing to reduce noise. The method is fast and accurate enough to filter ``easy'' pornographic images before a more robust identification process. Basically, it is a threshold used in the Hue component to select pixels related to skin. Then, an edge density map is computed to remove incorrectly detected regions. Assuming that the density of edges is low in skin regions, the edge pixels that have a high density are removed. Morphological operations are \red{also} used to reduce possible noise. Finally, mean and standard deviation of skin regions are calculated and another threshold is used to decide whether the image is pornographic or not.

\red{Zaidan et al.~\cite{Zaidan:2014} proposed an anti-pornography system based in two different stages: (i) skin detection and (ii) pornography classification. In the first stage, to detect skin regions from the image they combined the Bayesian method, a grouping histogram technique and back-propagation neural network based on the YCbCr and RGB color spaces. In the second stage, the features from the skin are extracted and classified, which determines if an image is pornographic.}

Although there are many approaches based on skin color detection to classify pornographic content, these works often have the disadvantage of a high rate of false positives since not all images with large areas of skin exposure are necessarily pornographic (\red{for example}, pictures with people wearing swimsuits, or sports-related images). Furthermore, another hurdle is the diversity of human skin color, making the classification process even more complicated. Another issue is that grayscale images cannot be classified using color related features~\cite{Ries:2012}. Shape-based approaches present the same problem, since they also make use of skin color information. It is important to mention that these are image-based approaches, and to be applied to video, a voting scheme should be considered.


\subsection{Local Features and Bag-of-Words Based Approaches} \label{subsec:localfeatures}

Another widely used approach in the literature of pornography detection is the employment of local feature extraction. Most of the works \red{applied} BoW models as an intermediate representation to encode the extracted local features followed by a classification step.

Deselaers et al.~\cite{Deselaers:2008} were the first to use local features with BoW models to classify pornographic content. They proposed an approach to filter and classify pornography in different categories. The SIFT detector was used to detect interesting points. They did not use any descriptor for describing the detected regions, claiming the advantage of patches as they provide color information. Next, each patch was reduced using Principal Component Analysis (PCA) and SVM was used as the final classifier. In a similar way, Lopes et al.~\cite{Lopes:2009a} performed image classification using the SIFT detector, HueSIFT descriptors and SVM classifier. A comparison is also made between the SIFT descriptors and HueSIFT applied to~pornography, showing that the combination of color information and local features perform better. In~\cite{Lopes:2009b}, the same authors extended their work to detect nudity in videos. To that end, they performed the same approach for selected video frames, but a majority voting scheme is held over the frames to set the video final classification.

To pre-filter pornographic images for isolating features of interest, Steel~\cite{Steel:2012} proposed a variant of the SIFT descriptor (Mask-SIFT) that uses a Gaussian pre-filter to remove all pixels of an image which are not related to skin. The image is then processed using a median filter to fill missing pixels and eliminate noise by creating a ``mask image''. Once this ``mask image'' is created, the SIFT descriptor is used to extract features from human related areas. For classification, the author developed a cascade based classifier filtering the images based on skin, shape and local features to determine whether an image is pornographic.

A study regarding the impact of moving patterns was made by Souza et al.~\cite{Souza:2012}. Color information is incorporated into the space-time interest point detector (STIP). They incorporated color information using the normalized-RGB color system at the detection phase called ColorSTIP. The local space-time regions are described in terms of histograms of Hue and combined with the default HOG-HOF feature histogram used by STIP, which they named as HueSTIP. BoW models are applied to encode the spatial-temporal features and SVM as final classifier.

Avila et al.~\cite{Avila:2013a} also presented an approach for pornographic video classification. First, the videos are segmented into shots and then the central frame of each shot is taken as keyframes to represent the video. After that, HueSIFT descriptors are extracted on a dense spatial grid. However, different from previous approaches, they used a BoW model extension --- the BossaNova mid-level image representation ---  to encode the local features. A SVM classifier is then applied to classify the keyframes extracted from each video shot and a majority voting scheme is employed to predict the video class.

With the aim of a fast feature detection and extraction, Zhuo et al.~\cite{Zhuo:2015} proposed a pornographic image recognition method based on the ORB binary descriptor. Their approach is divided in two parts: coarse detection and fine detection. The coarse detection identifies the non-pornographic images using a skin color detector conducted in YCbCr color space. For the remaining images containing much more skin-color regions, a fine detection is conducted. To that end, ORB descriptors are extracted from the skin-color regions and encoded by a BoW model. After that, they combined it with a 72-dimensional HSV color histogram of the whole image and trained a classification model using SVM for image recognition. Another fast approach based on binary descriptors was proposed by Yaghoubyan et al.~\cite{Yaghoubyan:2015}. First, they detected a region of interest (ROI) per input image by a threshold according to the ratio of skin color area in the SKN color space. Images with higher probability proceed to the further stages, otherwise they are labeled as non-pornographic. Then, keypoints are detected in the ROI eliminating noisy keypoints placed in the image background. FREAK binary descriptors are encoded by a BoW \red{representation} and a SVM classifier is applied to recognition. We also explored in~\cite{Caetano:2013} the benefit of \red{using} several binary descriptors and mid-level representation for videos to identify pornographic content with a SVM classifier and a majority voting scheme.

In contrast to previous approaches, which \red{employed} descriptors as local features, Ulges and Stahl~\cite{Ulges:2011} used the low frequency coefficients of the Discrete Cosine Transform (DCT) in YUV color space. 
As in previous approaches, BoW \red{technique} is also used as well as SVM for classification. To compare, the authors also implemented an approach based on color similar to Jones and Rehg~\cite{Jones:2002} and concluded that the \red{BoW-based} approach outperforms approaches based only on color. \red{Zhang et al.~\cite{Zhang:2013} employed visual attention model to find regions of interest (ROI) composed by pornographic content. Given an image, a face detection algorithm is applied to remove the face or ID photo from some non-pornographic images. Then, a visual salient map in compressed domain is computed to construct visual attention model, according to the large number of exposed skin regions. After, four features of color, texture, intensity and skin are extracted from the pornographic regions and a BoW model is applied. Finally, SVM is employed for classification.} 

In general, approaches based on local features have shown more satisfactory results than the ones based on color information~\cite{Ries:2012}. An important advantage of using local features is that they can be computed independent of color information. Moreover, another advantage is that these models compact the regions of the image in a fixed-size vector, making easy the comparison of image regions and/or the image as a whole. However, extraction of local features may prove to be more time-consuming than to examine image characteristics related to color.

	\section{BossaNova Video Descriptor (BNVD)}\label{proposal}

\newcommand\f{{\mathbf{O}}}
\newcommand\teste{{\mathbf{P}}}
\newcommand\h{{\mathbf{W}}}
\newcommand\vd{{\mathbf{vd}}}

Multimedia understanding is related to visual recognition in which we are interested in identifying, for example, pornography content and action recognition, in multimedia collections. Usually, the approaches to cope with those problems consider (i) extraction of local image descriptors; (ii) encoding of the local features in a mid-level representation; and (iii) classification and/or search of the image descriptor.  When a static image is considered, the visual recognition is based on, for instance, classification of a mid-level representation computed from the local image descriptors. A simple extension of this approach for video classification is to compute a mid-level representation by extracting local image descriptor from the video frames. Unfortunately, this simple and naive approach presents poor performance since it is susceptible to noise and it is not so discriminative for representing the video content.

To bypass this aforementioned problem, some methods are based on majority voting in which a binary classification is performed over the images. In a majority voting scheme to decide about the classes A and B, for example, the video is classified as belonging to the class~A if the number of images classified as class~A is greater than the number of images classified as class B. In literature, this approach is used in several applications, including pornography detection~\citep{Lopes:2009b,Avila:2013a, Caetano:2013}. Despite the good performance achieved on those works, some issues must be considered: (i) if a video contains few frames of a class~A, the video will be classified as class~B (as illustrated in Figure~\ref{fig:majorityvoting}), however in some applications such as video pornography, the existence of a few pornography frames must characterize the video as pornography; (ii) the presence of noise could exert influence over the number of frames of a specified class; and (iii) the probability of correct and wrong classifications are ignored since the evaluation is done by a static frame.

\begin{figure}[tb]
	\centering
	\includegraphics[width=1.0\linewidth]{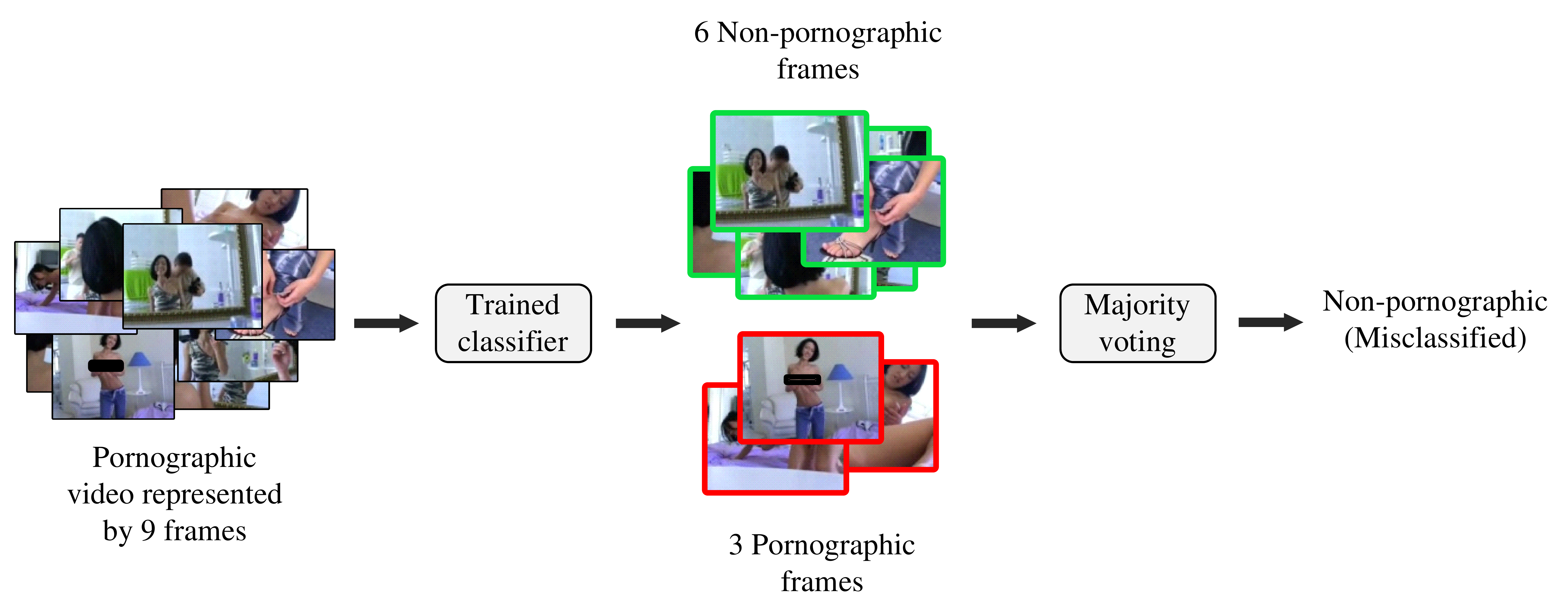}
	\caption{Example of video containing \red{nine} frames, \red{six} non-pornographic and \red{three} pornographic. The video will be misclassified to non-pornographic by the majority voting scheme, even been pornographic.}
	\label{fig:majorityvoting}
\end{figure}

To cope with these issues, the classification of a video must be done from a descriptor which represents all video content. In this way, we propose a strategy that aggregates the information of the mid-level representations of all video frames into a single representation. We present our video descriptor as follows.


%
%

Let ${\mathcal V}$ be a video sequence. ${\mathcal V} = \left\{f^i \right\}$, $i \in \left[1,N\right]$, where $f^i$ is the keyframe\footnote{A keyframe is a frame that represents the content of a logical unit, like a shot or scene, for example.} of the shot $i$ and $N$ is the number of keyframes. Let ${\mathcal Z} = \left\{\mathbf{z}^i \right\}$, $i \in \left[1,N\right]$  be a set of BossaNova vectors computed for the video ${\mathcal V}$ in which $\mathbf{z}^i$ is a BossaNova vector extracted for the keyframe $f^i$. Let $\f$ and $\teste$ be two functions for aggregating the information of BossaNova and the Bag of Visual Words. The BossaNova Video Descriptor (BNVD) can be modeled by a function $\h$ as follows:
\begin{eqnarray}\label{eq:desc:video}
\hspace{1cm} \f : \mathds{R}^B &\longrightarrow& \mathds{R}^B,  \nonumber \\
\teste : \mathds{R}^M &\longrightarrow& \mathds{R}^M,  \nonumber \\
\h : \mathds{R}^Z &\longrightarrow& \mathds{R}^Z,  \nonumber \\
    {\mathcal Z} &\longrightarrow& \h(\{\mathbf{z}^i\})= \left[ \left[o_{m,b}\right],p_m\right]^\text{T}, \nonumber\\ %
    o_{m,b} &=&\operatorname{\f} (\{z_{m,b}^i\}), \nonumber \\
    p_m &=&\operatorname{\teste} (\{t_{m}^i\}),
\end{eqnarray}
\vspace{0.025cm}
\noindent where $Z\subset\{1,...,M\} \times \{1,...,B\}$, and $\mathbf{z}^i = \left[ \left[z_{m,b}^i\right], t_m^i\right]^\text{T}$.


Intuitively, this new video descriptor represents a relation for each codeword to the codebook, since each BossaNova representation contains information regarding the dis\-tance-to-codeword distribution. The main goal of applying the functions $\f$ and $\teste$ to the BossaNova vectors is to employ a filtered-like operation to the entire video content which is represented by this mid-level representation. In this work, we study the behavior of our video descriptor by using the following functions: \emph{median}, \emph{mean}, \emph{min} and \emph{max}.

We also propose another video descriptor by using the classical BoW representation, called Bag-of-(Visual)-Words Video Descriptor (BoW-VD), that can be seen as a simplification of the BNVD since the distance to the codewords \red{($z_m$)} are ignored.


Figure~\ref{fig:majorityvoting} presents an example of applying \emph{min} and \emph{max} functions to the combination of three image descriptors. In this case, the \emph{min} function is applied to the local histograms $z_m$ (represented by the colored bins) and the \emph{max} function is applied to $t_m$ (represented by the dashed bin).


\begin{figure}[tb]
	\centering
	\includegraphics[width=1.0\linewidth]{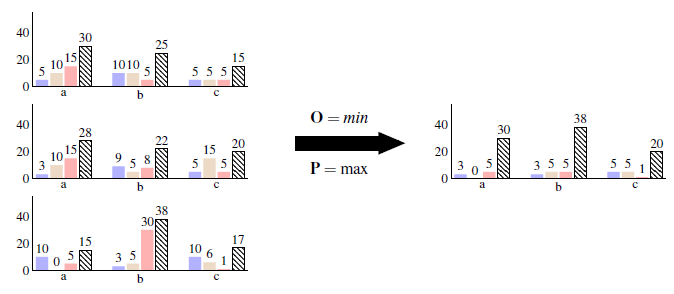}
	\caption{An example of computing the video descriptor from three image descriptors. The functions \emph{min} and \emph{max} are used for aggregating the set of BossaNovas.}
	\label{fig:majorityvoting}
\end{figure}


	\section{Experimental Results}\label{experiments}

\begin{figure*}[tb]
	\centering
	\includegraphics[width=0.85\linewidth]{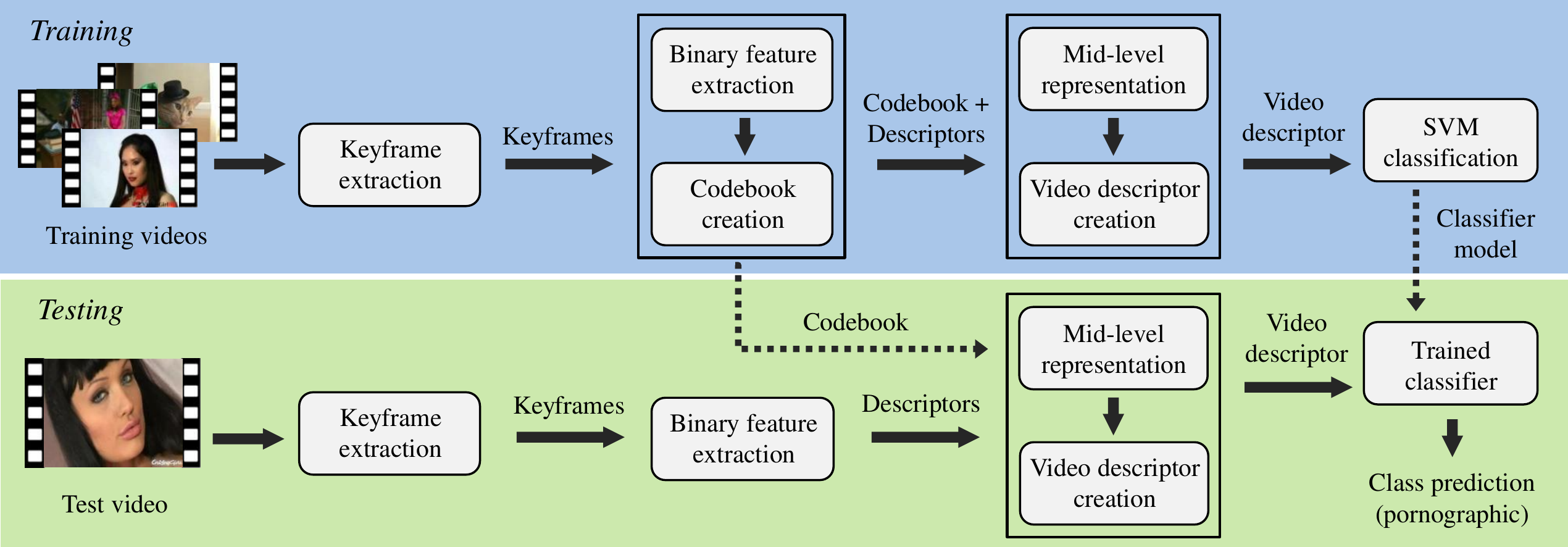}
	\caption{Overview of our methodology for pornography detection with the proposed video descriptor.}
	\label{fig:flow}
\end{figure*}

In this section, we present the results obtained through the evaluation of the proposed video descriptors applied to the pornography detection task. After describing our experimental setup (Section~\ref{protocol}) and giving details regarding Pornography dataset \citep{Avila:2013a} (Section~\ref{dataset}), we report the results achieved using the proposed video descriptors. Those results are organized in two groups: (i) quantitative (Section~\ref{evalandcomp}) and (ii) comparative analysis (Section~\ref{comparative}).

\subsection{Experimental Setup}\label{protocol}

In Figure~\ref{fig:flow}, we illustrate the overview of our experimental approach. Note that this methodology is adapted from \citep{Avila:2013a,Caetano:2013}, in which the binary image descriptors are computed followed by the computation of a mid-level representation for each keyframe. Our proposed video descriptors are computed according to Equation~\ref{eq:desc:video} for representing each video that will be classified.

For comparison purposes, we applied the same experimental setup described in \citep{Caetano:2013} to the binary feature extraction.
We extracted the most representative binary descriptors --- BRIEF~\cite{BRIEF}, ORB~\cite{ORB}, BRISK~\cite{BRISK}, FREAK~\cite{FREAK} and BinBoost~\cite{BinBoost} --- from patches of 16$\times$16~pixels sampled regularly using a step size of 6~pixels.
For BRIEF, ORB, BRISK, and FREAK, we obtained the implementations from OpenCV repository~\citep{OpenCV}; for BinBoost, we used the code available on the authors website\footnote{\url{http://www.cvlab.epfl.ch/research/detect/binboost}} and we assessed the length of the descriptor  which may vary among 8, 16 and 32 bytes.

Codebook learning is performed by a $k$-medians clustering algorithm with Hamming distance over one million randomly sampled descriptors.
The motivation to use such clustering technique is the fact that it is able to deal with the binary string nature of the binary descriptors.

For BossaNova mid-level feature extraction, we kept the parameter values the same as in \citep{Avila:2013a,Caetano:2013}: $B = $~10, $\lambda_{min} = $~0, $\lambda_{max} = $~3 and $s = $~10$^{-3}$, except for the number of visual codewords $M$. For comparison, we also extracted BoW mid-level features, obtained with hard-assignment coding and average pooling.

For binary classification (pornography vs. non-pornography), we used support vector machines (SVM) with a nonlinear kernel.
Kernel matrices are computed as $exp(-\gamma d(x, x'))$ in which $d$ is the distance and $\gamma$~is set to the inverse of the pairwise mean distances. The SVM  penalty parameter~$C$ is set to~10, which works best for the Pornography dataset.

All experiments were conducted on a 64-bit Ubuntu Linux machine powered by Intel$\circledR$ Xeon$\circledR$ CPU X5670 @ 2.93 GHz with 24~cores and 70~GB RAM.

\subsection{Pornography Dataset}\label{dataset}


We evaluate our approach on the challenging Pornography dataset \citep{Avila:2013a}. It is composed of 400~pornographic and 400~non-pornographic videos resulting in nearly 80~hours. The non-pornographic class is divided into two sub-classes: (i) ``easy'' with 200 randomly selected videos from the Internet; and (ii) ``difficult'', with 200 videos selected from text search queries as ``beach'', ``wrestling'' and ``swimming'', which is a group of high skin exposure becoming a quite challenging dataset. The video shots are already separated and a keyframe is selected to summarize the content of each shot into a static image (16,727~vi\-deo keyframes). The experimental protocol of the dataset consists of a classical 5-fold cross-validati\-on (640~videos for training and 160~for testing, on each fold). The video classification performance is reported by accuracy rate, where the final video label is obtained by the classification of the video descriptor, as described in Section~\ref{proposal}.

\begin{figure}[tb]
	\centering
	\includegraphics[width=\linewidth]{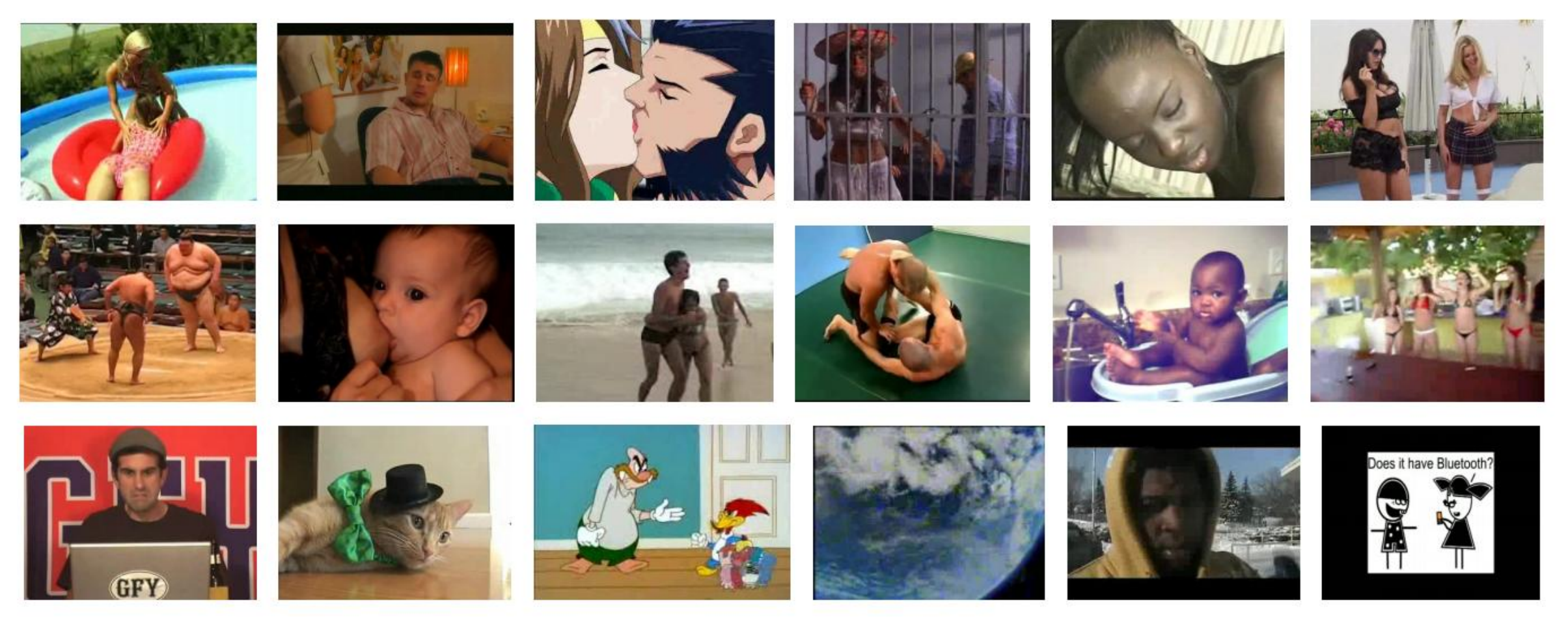}
	\caption{Example frames from the Pornography dataset \citep{Avila:2013a}, illustrating the diversity of pornographic videos (top row) and the non-pornographic videos (difficult on the middle row, a group with high skin exposure that is quite challenging for the detector, and easy at the bottom row).}
	\label{fig:pornography}
\end{figure}

Figure~\ref{fig:pornography} depicts the diversity of pornographic videos and the non-pornographic videos (difficult and easy).
The dataset is available upon request and the sign of a license agreement\footnote{\url{https://sites.google.com/site/pornographydatabase/}}.

\subsection{Quantitative Analysis}\label{evalandcomp}

In this section, we present discussions involving different aggregation functions used for combining the mid-level representation. We also evaluate the impact of codebook size on the proposed video descriptors.
Furthermore, we present the computational cost of different strategies to represent the data.

\begin{table*}[t!]
\begin{small}
	\begin{center}
		\caption{Video classification (\%) results (and standard deviations) of the proposed video descriptors, using Bag-of-Words (BoW), BossaNova, and different aggregation functions on Pornography dataset. For each method, we extracted a codebook size of $M = $ 256, as suggested in~\citep{Avila:2013a,Caetano:2013}. \vspace{0.1cm}}
		\begin{tabular}{lcccc}
			\toprule
			\textbf{Approach} & \textbf{\emph{max}} & \textbf{\emph{min}} & \textbf{\emph{mean}} & \textbf{\emph{median}} \\
			\toprule
			BoW-VD (BRIEF) & 85.55 $\pm$ 4 & 68.94 $\pm$ 5 & 86.79 $\pm$ 2 & \textbf{87.16} $\pm$ 2 \\
			BoW-VD (ORB) & 87.54 $\pm$ 3 & 77.18 $\pm$ 3 & 87.29 $\pm$ 2 & \textbf{88.04} $\pm$ 2  \\
			BoW-VD (BRISK) & 86.55 $\pm$ 2 & 71.70 $\pm$ 5 & \textbf{88.04} $\pm$ 3 & 87.04 $\pm$ 3  \\
			BoW-VD (FREAK) & 85.91 $\pm$ 4 & 80.30 $\pm$ 3 & \textbf{86.55} $\pm$ 3 & 86.42 $\pm$ 3 \\
			BoW-VD (BinBoost $d = $ 8) & 86.66 $\pm$ 2 & 72.44 $\pm$ 3 & 86.78 $\pm$ 1 & \textbf{87.78} $\pm$ 1 \\
			BoW-VD (BinBoost $d = $ 16) & 87.54 $\pm$ 2 & 73.32 $\pm$ 3 & \textbf{87.78} $\pm$ 1 & 87.77 $\pm$ 3  \\
			BoW-VD (BinBoost $d = $ 32) & 87.28 $\pm$ 2 & 72.82 $\pm$ 4 & 88.28 $\pm$ 2 & \textbf{88.53} $\pm$ 2  \\
			\midrule
			BNVD (BRIEF) & 88.16 $\pm$ 3 & 80.92 $\pm$ 4 & 88.66 $\pm$ 2 & \textbf{89.03} $\pm$ 1 \\
			BNVD (ORB) & 89.28 $\pm$ 3 & 84.42 $\pm$ 3 & \textbf{89.28} $\pm$ 3 & 89.02 $\pm$ 1  \\
			BNVD (BRISK) & 87.16 $\pm$ 2 & 82.30 $\pm$ 4 & 88.66 $\pm$ 0.5 & \textbf{89.27} $\pm$ 1  \\
			BNVD (FREAK) & 87.66 $\pm$ 2 & 85.66 $\pm$ 1 & 88.04 $\pm$ 1 & \textbf{89.66} $\pm$ 2  \\
			BNVD (BinBoost $d = $ 8) & 89.28 $\pm$ 3 & 82.05 $\pm$ 3 & 87.66 $\pm$ 1 & \textbf{90.77} $\pm$ 2  \\
			BNVD (BinBoost $d = $ 16) & 87.79 $\pm$ 2 & 82.42 $\pm$ 3 & 88.54 $\pm$ 2 & \textbf{90.90} $\pm$ 1  \\
			BNVD (BinBoost $d = $ 32) & 88.28 $\pm$ 2 & 81.92 $\pm$ 2 & 88.15 $\pm$ 1 & \textbf{89.41} $\pm$ 2 \\
			\bottomrule
		\end{tabular}%
		\label{tab:comparacaoFuncoes}%
	\end{center}
\end{small}
\end{table*}%

\subsubsection{Aggregation Function}

To compare different aggregation functions for the proposed video descriptors (as mentioned in Section~\ref{proposal}), we applied four different functions to $z_{m,b}^i$ and $t_{m}^i$: (i) \emph{max}, which selects the maximum existing value; (ii) \emph{min}, as the opposite of \emph{max}, selects the minimum existing value; (iii) \emph{mean}, which computes the average values; and (iv) \emph{median}, which selects the median values. Table~\ref{tab:comparacaoFuncoes} presents these experiments for both BossaNova and BoW. It can be seen that the \emph{median} function outperforms the others, except for the approach using the ORB descriptor for BNVD and BRISK, FREAK and BinBoost ($d = $ 16) for BoW. In such cases, \emph{mean} function presented a slightly higher accuracy value, but trebling the standard deviation on BNVD when compared to the \emph{median} function. Since the \emph{median} function presented the best results when compared to the other functions, from now on we use the results obtained by it as the main comparative method.

\subsubsection{Codebook Size}

\begin{figure}[ht!]
	\centering
	\subfloat[BoW Video Descriptor (BoW-VD)]{
	\includegraphics[width=1.0\linewidth]{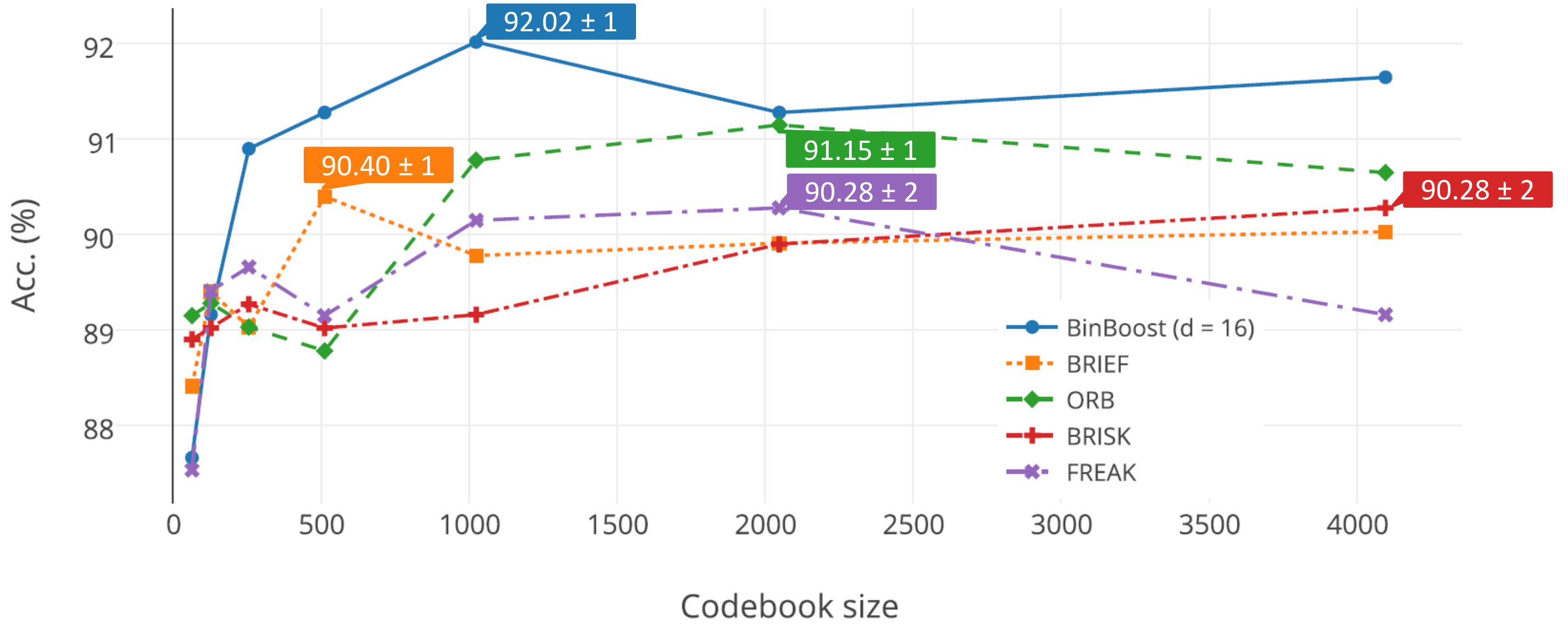}} \\
    \medskip \medskip
	\subfloat[BossaNova Video Descriptor (BNVD)]{
	\includegraphics[width=1.0\linewidth]{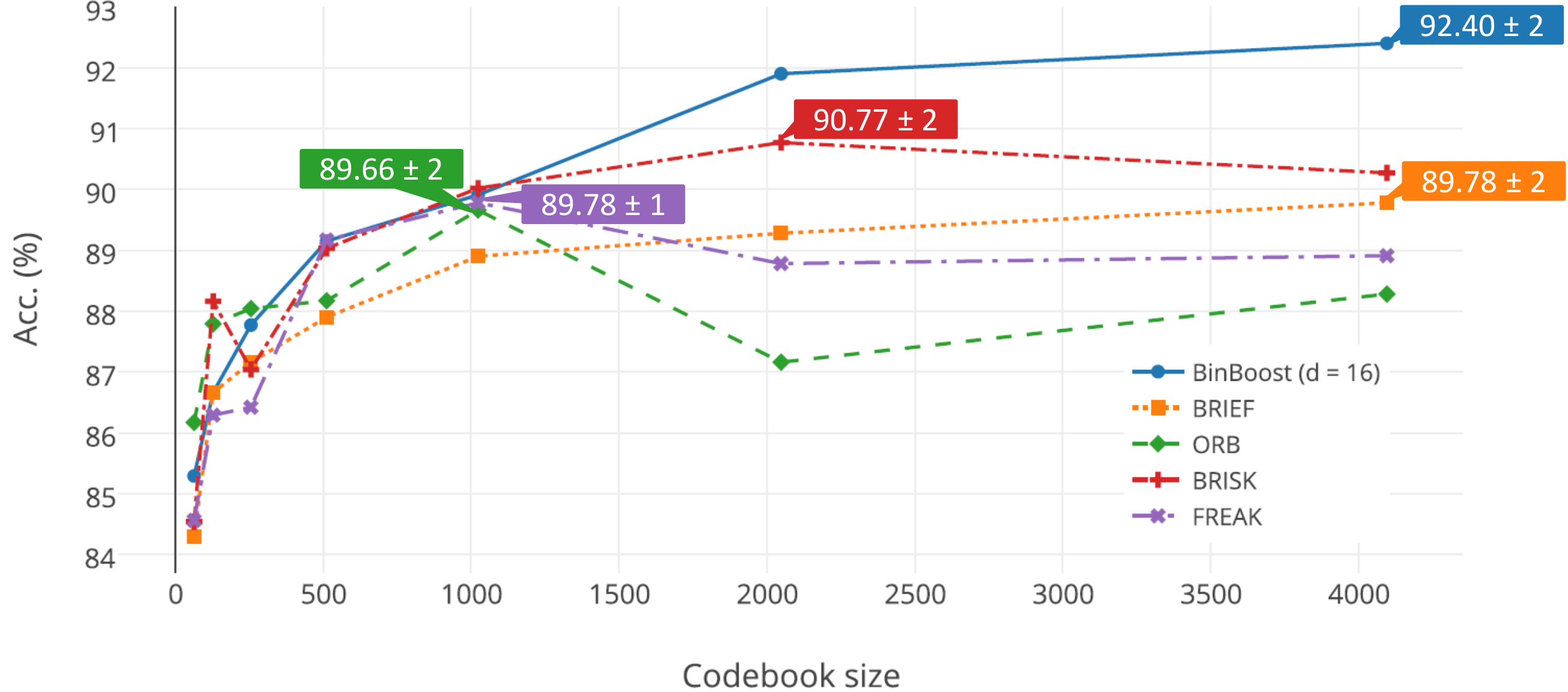}
	}
	\caption{Video classification (\%) results (and standard deviations) of video descriptors varying the codebook size $M$ on Pornography dataset.}
	\label{fig:plotCodebook}
\end{figure}

To provide a more comprehensive analysis of the proposed video descriptors, we evaluated their behavior by varying the codebook size $M$ in the classification performance. These experiments are shown in Figure~\ref{fig:plotCodebook}. We can observe that the approach using BinBoost ($d = $ 16) for both BoW-VD and BNVD achieved 92.40\% of accuracy (with a codebook size $M= $ 4096) and 92.02\% of accuracy (with a codebook size $M= $ 1024), respectively. 
The results obtained correspond to the mid-level parameters used from \citep{Avila:2013a}, so these parameters were not \red{tuned} for the proposed descriptors. Therefore, our approaches can achieve further results, as can be seen in Figure~\ref{fig:plotCodebook}. 

\subsubsection{Computational Cost}

Table~\ref{tab:times} presents a computational average time required to (i) extract the descriptors, (ii) create the codebook, (iii) generate the mid-level representations, and (iv) compute the proposed video descriptors. We can see how fast binary descriptors are compared to the HueSIFT descriptor used by \citep{Avila:2013a}. BRIEF and ORB descriptors presented an average extraction 10 times faster than HueSIFT and BinBoost ($d = $ 32), which is the slowest binary descriptor, becoming twice as fast.

\begin{table*}[t!]
\begin{small}
	\begin{center}
		\caption{Computational time required to: (i) extract the descriptors, (ii) create the codebook, (iii) generate the mid-level representations, and (iv) compute the proposed video descriptor. \vspace{0.1cm}}
		\begin{tabular}{lcccccc}
			\toprule
			\textbf{Approach} & \textbf{Descriptor} & \textbf{Codebook} & \textbf{BoW} & \textbf{BossaNova} & \textbf{BoW-VD} & \textbf{BNVD}\\
			\toprule
			HueSIFT \citep{Avila:2013a} & $2.54$ & $\mathbf{1.61\times10^{3}}$ & 0.12 & $1.33$ & $-$ & $-$\\
			BRIEF & $0.24$ & $4.75\times10^{3}$ & 0.05 & $0.50$ & 5.88 & $15.26$\\
			ORB & $\mathbf{0.23}$ & $6.51\times10^{3}$ & 0.04 & $0.49$ & 5.66 & $15.84$\\
			BRISK & $0.64$ & $10.23\times10^{3}$ & 0.08 & $0.83$ & 14.74 & $30.20$\\
			FREAK & $0.31$ & $9.48\times10^{3}$ & 0.08 & $0.84$ & 7.96 & $23.70$\\
			BinBoost $d = 16$ & $0.81$ & $16.12\times10^{3}$ & 0.02 & $0.23$ & 17.12 & $21.44$\\
			\bottomrule
		\end{tabular}%
		\label{tab:times}%
	\end{center}
\end{small}
\end{table*}%

It is also possible to observe in Table~\ref{tab:times} how the computational times to create the codebook differs from our approaches to \citep{Avila:2013a}. In this evaluation, Avila et al.~\citep{Avila:2013a} approach showed the best time. This is partly because our approaches \red{employed the} $k$-medians algorithm while Avila et al.~\citep{Avila:2013a} used the $k$-means algorithm. However, it is important to emphasize that the codebook creation step is performed during the training phase, i.e., an offline phase. Moreover, Table~\ref{tab:times} presents the computational time for creating the proposed video descriptor. The calculation is done by adding the average time values of extracting the descriptors with the average time for creating the mid-level representation. This value is multiplied by the average number of keyframes per video. Finally, we sum up the combination time between the mid level representations. Considering a codebook size of $M =$ 256, on average, this time is 0.001 and 0.012 milliseconds for BoW-VD and BNVD, respectively.


\subsection{A Comparative Analysis}\label{comparative}

In this section, we perform a comparison with classical and state-of-the-art methods, including both experiments with methods we have reimplemented ourselves and published results reported in the literature.

\subsubsection{Comparison to Classical Methods}\label{comp_classical}

In Table~\ref{tab:tab2}, we present a comparison of our proposed approaches to three different methods: (i) a implemented classical BoW method; (ii) our previous work \citep{Caetano:2013}; and (iii) an approach using video global pooling. 
We can notice a considerable improvement achieved by our proposed video descriptor in both approaches, BoW-VD and BNVD. BoW-VD reached 88.0\% of accuracy with ORB descriptor outperforming the basic BoW (BRISK) approach. Moreover, BNVD reached 90.9\% of accuracy with BinBoost descriptor ($d = $ 16). The comparison with our previous work \citep{Caetano:2013} is particularly relevant because we employed the same binary descriptors (BRIEF, ORB, BRISK and FREAK with default parameters). We note an absolute improvement of (by up to) 2.8\% from BossaNova to BNVD\red{, reducing the classification error by more 21\%}. 

Moreover, in Table~\ref{tab:tab2}, we present a comparison of our proposed approaches with an approach using video global pooling, i.e., creating just one mid-level representation for the video using all local features extracted from the keyframes. It can be seen that both BNVD and BoW-VD offered superior accuracy values when compared to a simple approach such as the global pooling of all the local features.

\begin{table*}[t!]
\begin{small}
\begin{center}
	\caption{ Video classification (\%) results (and standard deviations) of our approaches, implemented classical BoW methods, our previous work \citep{Caetano:2013}, and global pooling results on Pornography dataset~\citep{Avila:2013a}. For each method, we extracted a codebook size of $M = $ 256, as suggested in~\citep{Avila:2013a,Caetano:2013}. \vspace{0.1cm}}
	\begin{tabular}{clclc}
		\toprule
		& \textbf{Approach} & \textbf{Acc. (\%)} \vspace{0.075cm} & \textbf{Approach} & \textbf{Acc. (\%)} \vspace{0.075cm}\\
		\toprule
		\multirow{5}{*}{\textbf{Implemented}} & BoW (HueSIFT) \citep{Avila:2013a} & 83.0 $\pm$ 3 & BossaNova (HueSIFT) \citep{Avila:2013a} & 89.5 $\pm$ 1 \\
		\multirow{5}{*}{\textbf{methods}} & BoW (BRIEF) & 85.0 $\pm$ 3 & BossaNova (BRIEF) \citep{Caetano:2013} & 86.3 $\pm$ 3 \\
		& BoW (ORB) & 85.8 $\pm$ 2 & BossaNova (ORB) \citep{Caetano:2013} & 86.5 $\pm$ 3 \\
		& BoW (BRISK) & 87.0 $\pm$ 1 & BossaNova (BRISK) \citep{Caetano:2013} & 88.6 $\pm$ 2 \\
		& BoW (FREAK) & 85.8 $\pm$ 3 & BossaNova (FREAK) \citep{Caetano:2013} & 86.9 $\pm$ 3 \\
		& BoW (BinBoost $d = $ 16) & 86.7 $\pm$ 3 & BossaNova (BinBoost $d = 16$) & 89.4 $\pm$ 5 \\
		\midrule
		\multirow{4}{*}{\textbf{Our}} & BoW-VD (BRIEF) & 87.2 $\pm$ 2 & BNVD (BRIEF) & 89.0 $\pm$ 1 \\
		\multirow{4}{*}{\textbf{results}} & BoW-VD (ORB) & \textbf{88.0} $\pm$ 2 & BNVD (ORB)   & 89.0 $\pm$ 1 \\
		& BoW-VD (BRISK) & 87.0 $\pm$ 3 & BNVD (BRISK) & 89.3 $\pm$ 1 \\
		& BoW-VD (FREAK) & 86.4 $\pm$ 3 & BNVD (FREAK) & 89.7 $\pm$ 2 \\
		& BoW-VD (BinBoost $d = $ 16) & 87.8 $\pm$ 3 & BNVD (BinBoost $d = 16$)   & \textbf{90.9} $\pm$ 1 \\
		\midrule
		\multirow{4}{*}{\textbf{Global}} & BoW (BRIEF) & 72.7 $\pm$ 1 & BossaNova (BRIEF) & 80.3 $\pm$ 3 \\
		\multirow{4}{*}{\textbf{pooling}} & BoW (ORB) & 72.2 $\pm$ 2 & BossaNova (ORB)  & 78.8 $\pm$ 3 \\
		& BoW (BRISK) & 73.1 $\pm$ 2 & BossaNova (BRISK) & 79.9 $\pm$ 2 \\
		& BoW (FREAK) & 73.3 $\pm$ 3 & BossaNova (FREAK) & 79.6 $\pm$ 2 \\
		& BoW (BinBoost $d = $ 16) & 71.7 $\pm$ 2 & BossaNova (BinBoost $d = 16$) & 77.3 $\pm$ 1 \\
		\bottomrule
	\end{tabular}%
	\label{tab:tab2}%
\end{center}
\end{small}
\end{table*}%

\subsubsection{Comparison to State-of-the-art Methods}\label{comp_soa}

In Table~\ref{tab:tab3}, we present our results in comparison to the previous published methods that were also assessed in the Pornography dataset \citep{Avila:2013a}.
We can notice the improvement obtained by our proposed video descriptor in both approaches, BoW-VD and BNVD. BoW-VD reached 92.4\% of accuracy with BinBoost descriptor ($d = $ 16) outperforming the basic BoW approaches from Souza et al.~\cite{Souza:2012} and Avila et al.~\cite{Avila:2013a}. Moreover, BNVD reached 92.0\% of accuracy with BinBoost descriptor ($d = $ 16) also outperforming the BossaNova approaches from Avila et al.~\cite{Avila:2013a} and Caetano et al.~\cite{Caetano:2013}. Further, our results outperformed all the published local feature approaches which, as far as we know, used HueSIFT~\citep{Avila:2013a} and spatio-temporal features~\cite{Souza:2012}.


Tables~\ref{tab:confMatColorSTIP}--\ref{tab:confMatBNVD} show the confusion matrices for Souza et al.~\cite{Souza:2012}, Avila et al.~\cite{Avila:2013a}, BoW-VD (BinBoost $d = $ 16) and BNVD (BinBoost $d = $ 16), respectively. We can notice a higher value of true positives achieved by BNVD and BoW-VD (92.3\% and 93.0\%, respectively) against Souza et al.~BoW~(ColorSTIP)~\cite{Souza:2012} and Avila et al.~BossaNova~(HueSIFT)~\cite{Avila:2013a} (90.0\% and 88.2\%, respectively). Regarding the true negative values, our approaches achieved better when compared to Avila et al.~\cite{Avila:2013a} (91.8\% against 90.8\%) and a slightly close value when compared to Souza et al.~\cite{Souza:2012} (91.8\% against 92.0\%).

Table~\ref{tab:tab3} also presents results for Moustafa et al.~\citep{Moustafa:2015} convolutional neural network (CNN) approach. We note that, while we do not have greater accuracy than \red{method proposed by Moustafa et al.~\citep{Moustafa:2015}},  we remain close. Despite this, it is important to mention that CNN approaches can be computationally expensive requiring much cost in time and computational resources. On the other hand, our approach is much more simple employing a low-complexity alternative for feature extraction using binary descriptors and linear aggregation functions to BoW models.

In Figure~\ref{fig:roc}, we illustrate the ROC curves and the area under curve (AUC) for the Pornography dataset. Since 5-fold cross-validation protocol was applied, Figures~\ref{fig:roc} (a) and (b) presents the ROC curves for each fold and the mean curve for BoW-VD (BinBoost $d = $ 16) and BNVD (BinBoost $d = $ 16), respectively. Figure~\ref{fig:roc} (c) presents curves for our approaches in comparison to Avila et al.~\cite{Avila:2013a} and Caetano et al.~\cite{Caetano:2013}. Although the methods presented very similar curves, we can see that our proposed approaches achieved better AUC values. \emph{To the best of our knowledge, ours is the best result reported to date on Pornography dataset employing local feature descriptors}.

\begin{figure}[t!]
\begin{tabular}{cc}
\centering
\pgfdeclarelayer{front}
\pgfdeclarelayer{back}
\pgfsetlayers{back,main,front}

\tikzsetnextfilename{roc-bnvd}

\pgfplotstableread{source-figures/BNVD-ROC-fold0.txt}
\bnvdfoldone%
\pgfplotstableread{source-figures/BNVD-ROC-fold1.txt}
\bnvdfoldtwo%
\pgfplotstableread{source-figures/BNVD-ROC-fold2.txt}
\bnvdfoldthree%
\pgfplotstableread{source-figures/BNVD-ROC-fold3.txt}
\bnvdfoldfour%
\pgfplotstableread{source-figures/BNVD-ROC-fold4.txt}
\bnvdfoldfive%
\pgfplotstableread{source-figures/BNVD-ROC-mean.txt}
\bnvdfoldmean%

\resizebox{0.45\textwidth}{!} {
\begin{tikzpicture}
	\begin{axis}[
	xmax=1.0,xmin=0,
    ymin=0,ymax=1.0,
	legend style={at={(1,0)},anchor=south east},
	grid style={dotted,gray!90},
	grid=both,
	xlabel={False positive rate},
	ylabel={True positive rate}]
    \begin{pgfonlayer}{front}
	    \addplot [blue, draw, thick, mark=*,each nth point=20] table [x index=0, y index=1]{\bnvdfoldone};
	    \addlegendentry{ROC fold 0 (AUC = 0.969)}
	    \addplot [red, draw, thick, mark=square*, each nth point=18] table [x index=0, y index=1]{\bnvdfoldtwo};
	    \addlegendentry{ROC fold 1 (AUC = 0.958)}
	    \addplot [brown, draw, thick, mark=x, each nth point=20,mark size=4.0pt] table [x index=0, y index=1]{\bnvdfoldthree};
	    \addlegendentry{ROC fold 2 (AUC = 0.986)}
	    \addplot [green, draw, thick, mark=triangle*, each nth point=20] table [x index=0, y index=1]{\bnvdfoldfour};
	    \addlegendentry{ROC fold 3 (AUC = 0.970)}
	    \addplot [cyan, draw, thick, mark=diamond*, each nth point=20] table [x index=0, y index=1]{\bnvdfoldfive};
	    \addlegendentry{ROC fold 4 (AUC = 0.983)}
	    \addplot [black,dotted,draw, ultra thick, mark=star, each nth point=50] table [x index=0, y index=1]{\bnvdfoldmean};
	    \addlegendentry{Mean ROC (AUC = 0.973)}
    \end{pgfonlayer}
	\end{axis}
\end{tikzpicture}
} & \\
\hspace{1cm}\footnotesize{(a) ROC curves for BNVD (BinBoost $d =$ 16)} \vspace{0.15cm} \\
\pgfdeclarelayer{front}
\pgfdeclarelayer{back}
\pgfsetlayers{back,main,front}

\tikzsetnextfilename{roc-bowvd}

\pgfplotstableread{source-figures/BoW-VD-ROC-fold0.txt}
\bnvdfoldone%
\pgfplotstableread{source-figures/BoW-VD-ROC-fold1.txt}
\bnvdfoldtwo%
\pgfplotstableread{source-figures/BoW-VD-ROC-fold2.txt}
\bnvdfoldthree%
\pgfplotstableread{source-figures/BoW-VD-ROC-fold3.txt}
\bnvdfoldfour%
\pgfplotstableread{source-figures/BoW-VD-ROC-fold4.txt}
\bnvdfoldfive%
\pgfplotstableread{source-figures/BoW-VD-ROC-mean.txt}
\bnvdfoldmean%

\resizebox{0.45\textwidth}{!} {
\begin{tikzpicture}
	\begin{axis}[
	xmax=1.0,xmin=0,
    ymin=0,ymax=1.0,
	legend style={at={(1,0)},anchor=south east},
	grid style={dotted,gray!90},
	grid=both,
	xlabel={False positive rate},
	ylabel={True positive rate}]
    \begin{pgfonlayer}{front}
	    \addplot [blue, draw, thick, mark=*,each nth point=20] table [x index=0, y index=1]{\bnvdfoldone};
	    \addlegendentry{ROC fold 0 (AUC = 0.969)}
	    \addplot [red, draw, thick, mark=square*, each nth point=20] table [x index=0, y index=1]{\bnvdfoldtwo};
	    \addlegendentry{ROC fold 1 (AUC = 0.965)}
	    \addplot [brown, draw, thick, mark=x, each nth point=20,mark size=4.0pt] table [x index=0, y index=1]{\bnvdfoldthree};
	    \addlegendentry{ROC fold 2 (AUC = 0.985)}
	    \addplot [green, draw, thick, mark=triangle*, each nth point=20] table [x index=0, y index=1]{\bnvdfoldfour};
	    \addlegendentry{ROC fold 3 (AUC = 0.974)}
	    \addplot [cyan, draw, thick, mark=diamond*, each nth point=20] table [x index=0, y index=1]{\bnvdfoldfive};
	    \addlegendentry{ROC fold 4 (AUC = 0.985)}
	    \addplot [black,dotted,draw, ultra thick, mark=star, each nth point=50] table [x index=0, y index=1]{\bnvdfoldmean};
	    \addlegendentry{Mean ROC (AUC = 0.976)}
     \end{pgfonlayer}
	\end{axis}
\end{tikzpicture}
} & \\
\hspace{1cm}\footnotesize{(b) ROC curves for BoW-VD (BinBoos $d =$ 16)} \vspace{0.15cm} \\
\pgfdeclarelayer{front}
\pgfdeclarelayer{back}
\pgfsetlayers{back,main,front}
\pgfplotstableread{source-figures/BNVD-ROC-mean.txt}
\bnvdfoldmean%
\pgfplotstableread{source-figures/BoW-VD-ROC-mean.txt}
\bowvdfoldmean%
\pgfplotstableread{source-figures/SampledBossaNova-BRISK-ROC-mean.txt}
\bnbriskfoldmean%
\pgfplotstableread{source-figures/SampledBossaNova-HueSIFT-ROC-mean.txt}
\bnhuesiftfoldmean%

\tikzsetnextfilename{roc-mean-all}

\resizebox{0.45\textwidth}{!} {
\begin{tikzpicture}
	\begin{axis}[
	xmax=1.0,xmin=0,
    ymin=0,ymax=1.0,
	legend style={at={(1,0)},anchor=south east},
	grid style={dotted,gray!90},
	grid=both,
	xlabel={False positive rate},
	ylabel={True positive rate}]
    \begin{pgfonlayer}{front}
	    \addplot [red, thick,mark=square*, each nth point=100] table [x index=0, y index=1]{\bnvdfoldmean};
	    \addlegendentry{BNVD (AUC = \textbf{0.973})}
	    \addplot [blue,draw,  thick,mark=*, each nth point=100] table [x index=0, y index=1]{\bowvdfoldmean};
	    \addlegendentry{BoW-VD (AUC = \textbf{0.976})}
	    \addplot [green,draw,  thick, mark=triangle*, ,mark size=4.0pt, each nth point=250] table [x index=0, y index=1]{\bnbriskfoldmean};
	    \addlegendentry{Caetano et al.  (AUC = 0.960)}
   	    \addplot [YellowOrange,draw,  thick, mark=star, ,mark size=4.0pt, each nth point=240] table [x index=0, y index=1]{\bnhuesiftfoldmean};
   	    \addlegendentry{Avila et al.  (AUC = 0.954)}
	\end{pgfonlayer}
	\end{axis}
\end{tikzpicture}
}
 & \\
\hspace{1cm}\footnotesize{(c) Mean ROC curves}\\
\end{tabular}
\caption{ROC curves and the AUC for the Pornography dataset.}
\label{fig:roc}
\end{figure}

\begin{table}[!t]
\begin{small}
	\centering
	\caption{Video classification (\%) results (and standard deviations) of our approaches and published results on Pornography dataset~\citep{Avila:2013a}. \vspace{0.1cm}}
	\begin{tabular}{clc}
		\toprule
		& \textbf{Approach} & \textbf{Acc. (\%)}\\
		\toprule
		& Souza et al. [BoW (STIP)] \citep{Souza:2012} & 89.6 $\pm$ $-$ \\
		& Souza et al. [BoW (HueSTIP)] \citep{Souza:2012} & 90.0 $\pm$ $-$ \\
		& Souza et al. [BoW (ColorSTIP)] \citep{Souza:2012} & 91.0 $\pm$ $-$ \\
		\textbf{Published} & Avila et al. [BoW (HueSIFT)] \citep{Avila:2013a} & 83.0 $\pm$ 3 \\
		\textbf{results} & Avila et al. [BOSSA (HueSIFT)] \citep{Avila:2011} & 87.1 $\pm$ 2 \\
		& Avila et al. [BossaNova (HueSIFT)] \citep{Avila:2013a} & 89.5 $\pm$ 1 \\
		& Caetano et al. [BossaNova (BRISK)] \citep{Caetano:2013} & 88.6 $\pm$ 2 \\
		& Moustafa et al. [CNN (AGbNet)] \citep{Moustafa:2015} & \textbf{94.1} $\pm$ 2 \\
		\midrule
		\textbf{Our} & BoW-VD (BinBoost $d = $ 16) & \textbf{92.4} $\pm$ 2 \\
		\textbf{results}& BNVD (BinBoost $d = 16$)   & \textbf{92.0} $\pm$ 1 \\
		\bottomrule
	\end{tabular}%
	\label{tab:tab3}%
\end{small}
\end{table}%

\begin{table}[t!]
\begin{small}
\begin{minipage}[b]{\linewidth}
	\centering
		\caption{Average confusion matrix for Souza et al. [BoW (ColorSTIP)]~\citep{Souza:2012} approach.}
		\begin{tabular}{r|r|c|c|}
			\multicolumn{1}{r}{} & \multicolumn{1}{r}{} & \multicolumn{2}{c}{\textbf{Video was labeled as}} \bigstrut[b]\\
			\cline{3-4}    \multicolumn{1}{r}{} & \multicolumn{1}{c|}{} & porn & non-porn \bigstrut\\
			\cline{2-4}    \textbf{Video} & \multicolumn{1}{c|}{porn} & 90.0\%     & 10.0\% \bigstrut\\
			\cline{2-4}    \textbf{class} & \multicolumn{1}{c|}{non-porn} & 8.0\%     & 92.0\% \bigstrut\\
			\cline{2-4}    \end{tabular}%
		\label{tab:confMatColorSTIP}%
\end{minipage}
\medskip \medskip \medskip

\begin{minipage}[b]{\linewidth}
	\centering
		\caption{Average confusion matrix for Avila et al. [BossaNova (HueSIFT)]~\citep{Avila:2013a} approach.}
		\begin{tabular}{r|r|c|c|}
			\multicolumn{1}{r}{} & \multicolumn{1}{r}{} & \multicolumn{2}{c}{\textbf{Video was labeled as}} \bigstrut[b]\\
			\cline{3-4}    \multicolumn{1}{r}{} & \multicolumn{1}{c|}{} & porn & non-porn \bigstrut\\
			\cline{2-4}    \textbf{Video} & \multicolumn{1}{c|}{porn} & 88.2\%     & 11.8\% \bigstrut\\
			\cline{2-4}    \textbf{class} & \multicolumn{1}{c|}{non-porn} & 9.2\%     & 90.8\% \bigstrut\\
			\cline{2-4}    \end{tabular}%
		\label{tab:confMatHueSIFT}%
\end{minipage}
\medskip \medskip

\begin{minipage}[b]{\linewidth}
	\centering
	\caption{Average confusion matrix for our BoW-VD (BinBoost $d = $ 16) approach.}
	\begin{tabular}{r|r|c|c|}
		\multicolumn{1}{r}{} & \multicolumn{1}{r}{} & \multicolumn{2}{c}{\textbf{Video was labeled as}} \bigstrut[b]\\
		\cline{3-4}    \multicolumn{1}{r}{} & \multicolumn{1}{c|}{} & porn & non-porn \bigstrut\\
		\cline{2-4}    \textbf{Video} & \multicolumn{1}{c|}{porn} & 93.0\%     & 7.0\% \bigstrut\\
		\cline{2-4}    \textbf{class} & \multicolumn{1}{c|}{non-porn} & 8.2\%     & 91.8\% \bigstrut\\
		\cline{2-4}    \end{tabular}%
	\label{tab:confMatBoW-VD}%
\end{minipage}
\medskip \medskip

\begin{minipage}[b]{\linewidth}
	\centering
	\caption{Average confusion matrix for our BNVD (BinBoost $d = 16$) approach.}
	\begin{tabular}{r|r|c|c|}
		\multicolumn{1}{r}{} & \multicolumn{1}{r}{} & \multicolumn{2}{c}{\textbf{Video was labeled as}} \bigstrut[b]\\
		\cline{3-4}    \multicolumn{1}{r}{} & \multicolumn{1}{c|}{} & porn & non-porn \bigstrut\\
		\cline{2-4}    \textbf{Video} & \multicolumn{1}{c|}{porn} & 92.3\%     & 7.7\% \bigstrut\\
		\cline{2-4}    \textbf{class} & \multicolumn{1}{c|}{non-porn} & 8.2\%     & 91.8\% \bigstrut\\
		\cline{2-4}    \end{tabular}%
	\label{tab:confMatBNVD}%
\end{minipage}
\end{small}
\end{table}%

\begin{figure}[]
\begin{center}
	\centering
		\includegraphics[width=2cm,height=1.5cm]{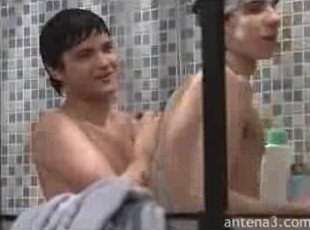}\hspace{.2cm}
		\includegraphics[width=2cm,height=1.5cm]{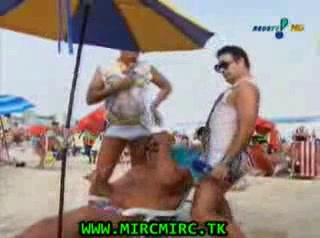}\hspace{.2cm}
		\includegraphics[width=2cm,height=1.5cm]{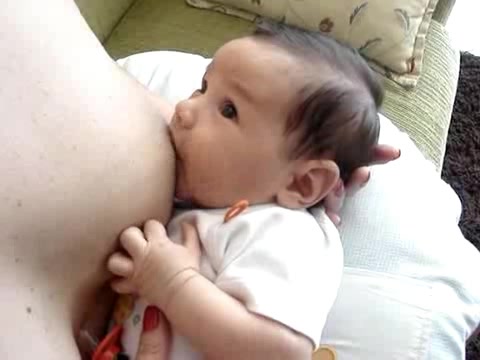} \\
	\centerline{\footnotesize (a) Misclassified non-pornographic videos examples.}\medskip \medskip
		\includegraphics[width=2cm,height=1.5cm]{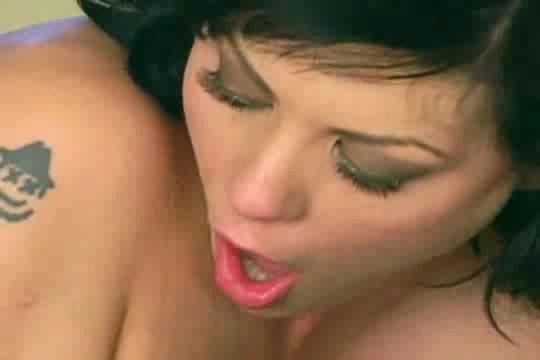}\hspace{.2cm}
		\includegraphics[width=2cm,height=1.5cm]{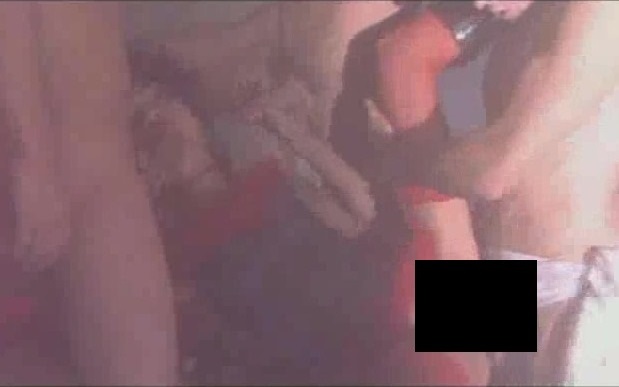}\hspace{.2cm}
		\includegraphics[width=2cm,height=1.5cm]{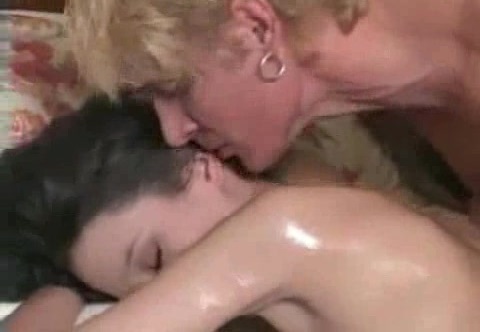}
	\centerline{\footnotesize (b) Misclassified pornographic videos examples. }
    \vspace{-.25cm}
	\caption{Cases where our method fails. (a) Misclassified non-pornographic videos. (b) Misclassified pornographic videos.}
	\label{fig:misclassifications}
\end{center}
\end{figure}

We also investigated the cases where our method failed. The misclassified non-pornographic videos correspond to very challenging cases, such as breastfeeding sequences, sequences of children being bathed, and beach scenes (as illustrated in Figure~\ref{fig:misclassifications}(a)). In addition, the analysis of the misclassified pornographic videos revealed that the method presented difficulties with poor quality videos or when the clip is borderline pornographic, with few explicit elements (as illustrated in Figure~\ref{fig:misclassifications}(b)). The same difficulty was also reported by Avila et al.~\citep{Avila:2013a}.

	\section{Conclusions}\label{conclusion}

The task of detecting and filtering pornographic visual content from the Internet is a well known concern in many environments, from schools to workplaces. In view of that, we presented a method for video pornography detection. The proposed method integrated the advantages of concepts presented in reference works in pornography detection area, contributing to improve the state of the art.

Specifically, our work focused on the supervised classification of pornographic \red{content in} videos based on local feature descriptors coded on mid-level representations. The description of the local features was performed using binary descriptors, a low complexity alternative, and a mid-level representation called BossaNova, an extension of the Bag-of-Words model that richly preserves the visual information.

Our approach is based on a combination of mid-level representations, yielding the Bag-of-Words Video Descriptor (BoW-VD) and BossaNova Video Descriptor (BNVD). The main goal is to apply aggregation functions to combine the collections of mid-level representations, thus creating a filtered-like operation on all the video content which are represented by this mid-level representation. The results validated the proposed video descriptors in the context of pornography detection, 
outperforming the state of the art by more than two percentage points, reducing the classification error by over 16\% (from 9.1\% to 7.6\%).

In order to provide more comprehensive analysis of our video descriptors, we propose evaluating their behavior on other video classification problems. 

	\section{Acknowledgments}
	The authors are thankful to CNPq, CAPES and FAPEMIG, Brazilian research and development agencies for the support to this work.
	
	\bibliographystyle{elsarticle-num}
	\bibliography{Bibliography}

\begin{thebibliography}{10}
\expandafter\ifx\csname url\endcsname\relax
  \def\url#1{\texttt{#1}}\fi
\expandafter\ifx\csname urlprefix\endcsname\relax\def\urlprefix{URL }\fi
\expandafter\ifx\csname href\endcsname\relax
  \def\href#1#2{#2} \def\path#1{#1}\fi

\bibitem{Short:2012}
M.~B. Short, L.~Black, A.~H. Smith, C.~T. Wetterneck, D.~E. Wells, A review of
  internet pornography use research: Methodology and content from the past 10
  years, Cyberpsychology, Behavior, and Social Networking 15~(1) (2012) 13--23.

\bibitem{Deselaers:2008}
T.~Deselaers, L.~Pimenidis, H.~Ney, Bag-of-visual-words models for adult image
  classification and filtering, in: International Conference on Pattern
  Recognition (ICPR), 2008, pp. 1--4.

\bibitem{Valle:2011}
E.~Valle, S.~Avila, F.~Souza, M.~Coelho, A.~de~A.~Ara{\'u}jo, Content-based
  filtering for video sharing social networks, in: Brazilian Symposium on
  Information and Computer System Security (SBSeg), 2012, pp. 625--638.

\bibitem{Avila:2013a}
S.~Avila, N.~Thome, M.~Cord, E.~Valle, A.~de~A.~Ara\'{u}jo, Pooling in image
  representation: the visual codeword point of view, Computer Vision and Image
  Understanding (CVIU) 117~(5) (2013) 453--465.

\bibitem{Lopes:2009a}
A.~Lopes, S.~Avila, A.~Peixoto, R.~Oliveira, A.~de~A.~Ara\'{u}jo, A
  bag-of-features approach based on {Hue-SIFT} descriptor for nude detection,
  in: European Signal Processing Conference (EUSIPCO), 2009, pp. 1552--1556.

\bibitem{Forsyth:1996}
D.~Forsyth, M.~Fleck, Identifying nude pictures, in: IEEE Workshop on
  Applications of Computer Vision (WACV), 1996, pp. 103--108.

\bibitem{Forsyth:1997}
D.~A. Forsyth, M.~M. Fleck, Body plans, in: IEEE Conference on Computer Vision
  and Pattern Recognition (CVPR), 1997, pp. 678--683.

\bibitem{Forsyth:1999}
D.~A. Forsyth, M.~M. Fleck, Automatic detection of human nudes, International
  Journal of Computer Vision (IJCV) 32~(1) (1999) 63--77.

\bibitem{Jones:2002}
M.~Jones, J.~Rehg, Statistical color models with application to skin detection,
  International Journal of Computer Vision (IJCV) 46~(1) (2002) 81--96.

\bibitem{Zheng:2004}
H.~Zheng, M.~Daoudi, B.~Jedynak, Blocking adult images based on statistical
  skin detection, Electronic Letters on Computer Vision and Image Analysis
  (ELCVIA) 4~(2) (2004) 1--14.

\bibitem{Rowley:2006}
H.~Rowley, Y.~Jing, S.~Baluja, {Large scale image-based adult-content
  filtering}, in: International Conference on Computer Vision Theory and
  Applications (VISAPP), 2006, pp. 290--296.

\bibitem{Hu:2007}
W.~Hu, O.~Wu, Z.~Chen, Z.~Fu, S.~Maybank, Recognition of pornographic web pages
  by classifying texts and images, IEEE Transactions on Pattern Analysis and
  Machine Intelligence (TPAMI) 29~(6) (2007) 1019--1034.

\bibitem{Lopes:2009b}
A.~Lopes, S.~Avila, A.~Peixoto, R.~Oliveira, M.~Coelho, A.~de~A.~Ara\'{u}jo,
  Nude detection in video using bag-of-visual-features, in: Brazilian Symposium
  on Computer Graphics and Image Processing (SIBGRAPI), 2009, pp. 224--231.
\newblock \href {http://dx.doi.org/10.1109/SIBGRAPI.2009.32}
  {\path{doi:10.1109/SIBGRAPI.2009.32}}.

\bibitem{Ulges:2011}
A.~Ulges, A.~Stahl, Automatic detection of child pornography using color visual
  words, in: IEEE International Conference on Multimedia and Expo (ICME), 2011,
  pp. 1--6.

\bibitem{Steel:2012}
C.~Steel, The mask-sift cascading classifier for pornography detection, in:
  World Congress on Internet Security (WorldCIS), 2012, pp. 139--142.

\bibitem{Yu:2014}
J.-J. Yu, S.-W. Han, Skin detection for adult image identification, in:
  International Conference on Advanced Communication Technology (ICACT), 2014,
  pp. 645--648.

\bibitem{Caetano:2013}
C.~Caetano, S.~Avila, S.~Guimar{\~a}es, A.~de~A.~Ara{\'u}jo, Representing local
  binary descriptors with {BossaNova} for visual recognition, in: Symposium On
  Applied Computing (ACM SAC), 2014, pp. 49--54.
\newblock \href {http://dx.doi.org/10.1145/2554850.2555058}
  {\path{doi:10.1145/2554850.2555058}}.

\bibitem{Zhuo:2015}
L.~Zhuo, Z.~Geng, J.~Zhang, X.~G. Li, {ORB} feature based web pornographic
  image recognition, Neurocomputing 173~(3) (2016) 511--517.

\bibitem{Ries:2012}
C.~Ries, R.~Lienhart, A survey on visual adult image recognition, Multimedia
  Tools and Applications (MTA) 69~(3) (2014) 661--688.

\bibitem{Agarwal:2004}
S.~Agarwal, A.~Awan, D.~Roth, Learning to detect objects in images via a
  sparse, part-based representation, IEEE Transactions on Pattern Analysis and
  Machine Intelligence (TPAMI) 26~(11) (2004) 1475--1490.

\bibitem{Yang:2007}
J.~Yang, Y.-G. Jiang, A.~G. Hauptmann, C.-W. Ngo, Evaluating
  bag-of-visual-words representations in scene classification, in:
  International Workshop on Workshop on Multimedia Information Retrieval (MIR),
  2007, pp. 197--206.

\bibitem{HueSIFT}
K.~E.~A. Van~de Sande, T.~Gevers, C.~G.~M. Snoek, Evaluating color descriptors
  for object and scene recognition, IEEE Transactions on Pattern Analysis and
  Machine Intelligence (TPAMI) 32~(9) (2010) 1582--1596.

\bibitem{Lowe:2004}
D.~Lowe, Distinctive image features from scale-invariant keypoints,
  International Journal of Computer Vision (IJCV) 60~(2) (2004) 91--110.

\bibitem{Caetano:2014:EUSIPCO}
C.~Caetano, S.~Avila, S.~Guimar{\~a}es, A.~de~A.~Ara{\'u}jo, Pornography
  detection using {BossaNova} video descriptor, in: European Signal Processing
  Conferente (EUSIPCO), 2014, pp. 1681--1685.

\bibitem{Chatfield:2011}
K.~Chatfield, V.~Lemtexpitsky, A.~Vedaldi, A.~Zisserman, {The devil is in the
  details: an evaluation of recent feature encoding methods}, in: British
  Machine Vision Conference (BMVC), 2011, pp. 1--12.

\bibitem{SURF}
H.~Bay, A.~Ess, T.~Tuytelaars, L.-V. Gool, Speeded-up robust features {(SURF)},
  Computer Vision and Image Understanding (CVIU) 110~(3) (2008) 346--359.

\bibitem{Sivic:2003}
J.~Sivic, A.~Zisserman, {Video Google}: a text retrieval approach to object
  matching in videos, in: International Conference on Computer Vision (ICCV),
  2003, pp. 1470--.

\bibitem{Datar:2004}
M.~Datar, N.~Immorlica, P.~Indyk, V.~S. Mirrokni, Locality-sensitive hashing
  scheme based on p-stable distributions, in: 20th Annual Symposium on
  Computational Geometry (SCG), 2004, pp. 253--262.

\bibitem{Gong:2011}
Y.~Gong, S.~Lazebnik, Iterative quantization: A procrustean approach to
  learning binary codes, in: IEEE Conference on Computer Vision and Pattern
  Recognition (CVPR), 2011, pp. 817--824.

\bibitem{Shen:2014}
F.~Shen, C.~Shen, Q.~Shi, A.~van~den Hengel, Z.~Tang, H.~T. Shen, Hashing on
  nonlinear manifolds, IEEE Transactions on Image Processing (TIP) 24~(6)
  (2015) 1839--1851.

\bibitem{Shen:2015}
F.~Shen, W.~Liu, S.~Zhang, Y.~Yang, H.~T. Shen, Learning binary codes for
  maximum inner product search, in: IEEE International Conference on Computer
  Vision (ICCV), 2015, pp. 4148--4156.

\bibitem{Santos:2015}
C.~E. dos Santos, E.~Kijak, G.~Gravier, W.~R. Schwartz, Learning to hash faces
  using large feature vectors, in: International Workshop on Content-Based
  Multimedia Indexing (CBMI), 2015, pp. 1--6.

\bibitem{GLOH}
K.~Mikolajczyk, C.~Schmid, A performance evaluation of local descriptors, IEEE
  Transactions on Pattern Analysis and Machine Intelligence (TPAMI) 27~(10)
  (2005) 1615--1630.

\bibitem{Canclini:2013}
A.~Canclini, M.~Cesana, R.~A., M.~Tagliasacchi, J.~Ascenso, C.~R., Evaluation
  of low-complexity visual feature detectors and descriptors, in: International
  Conference on Digital Signal Processing (DSP), 2013, pp. 1--7.

\bibitem{Morel:2011}
J.~Morel, G.~Yu, Is sift scale invariant?, Inverse Problems and Imaging 5~(1)
  (2011) 115--136.

\bibitem{BRIEF}
M.~Calonder, V.~Lepetit, C.~Strecha, P.~Fua, {BRIEF}: binary robust independent
  elementary features, in: European Conference on Computer Vision: Part IV
  (ECCV), 2010, pp. 778--792.

\bibitem{ORB}
E.~Rublee, V.~Rabaud, K.~Konolige, G.~Bradski, {ORB}: An efficient alternative
  to {SIFT} or {SURF}, in: International Conference on Computer Vision (ICCV),
  2011, pp. 2564--2571.

\bibitem{BRISK}
S.~Leutenegger, M.~Chli, R.~Siegwart, {BRISK}: Binary robust invariant scalable
  keypoints, in: International Conference on Computer Vision (ICCV), 2011, pp.
  2548--2555.

\bibitem{FREAK}
A.~Alahi, R.~Ortiz, P.~Vandergheynst, {FREAK}: Fast retina keypoint, in: IEEE
  Conference on Computer Vision and Pattern Recognition (CVPR), 2012, pp.
  510--517.

\bibitem{BinBoost}
V.~L. T.~Trzcinski, M.~Christoudias, P.~Fua, {Boosting Binary Keypoint
  Descriptors}, in: IEEE Conference on Computer Vision and Pattern Recognition
  (CVPR), 2013, pp. 2874--2881.

\bibitem{AdaBoost}
Y.~Freund, R.~E. Schapire, A decision-theoretic generalization of on-line
  learning and an application to boosting, Journal of Computer and System
  Sciences 55~(1) (1997) 119--139.

\bibitem{Avila:2011}
S.~Avila, N.~Thome, M.~Cord, E.~Valle, A.~de~A.~Ara\'{u}jo, {BOSSA}: Extended
  {BoW} formalism for image classification, in: International Conference on
  Image Processing (ICIP), 2011, pp. 2909--2912.

\bibitem{Sanchez:2013}
J.~S{\'a}nchez, F.~Perronnin, T.~Mensink, J.~Verbeek, Image classification with
  the fisher vector: Theory and practice, International Journal of Computer
  Vision (IJCV) 105~(3) (2013) 222--245.

\bibitem{Zhou:2010}
X.~Zhou, K.~Yu, T.~Zhang, T.~Huang, Image classification using super-vector
  coding of local image descriptors, in: European Conference on Computer Vision
  (ECCV), 2010, pp. 141--154.

\bibitem{Lee:2007}
J.-S. Lee, Y.-M. Kuo, P.-C. Chung, E.-L. Chen, Naked image detection based on
  adaptive and extensible skin color model, Pattern Recognition 40~(8) (2007)
  2261--2270.

\bibitem{Lee:2013}
J.-S. Lee, F.-S. Yu, K.-Y. Huang, Pornography detection based on morphological
  features, International Journal of Computer, Consumer and Control (IJ3C) 2
  (2013) 56--64.

\bibitem{Zaidan:2014}
A.~Zaidan, N.~Ahmad, H.~Karim, M.~Larbani, B.~Zaidan, A.~Sali, On the
  multi-agent learning neural and {Bayesian} methods in skin detector and
  pornography classifier: An automated anti-pornography system, Neurocomputing
  131~(5) (2014) 397--418.

\bibitem{Souza:2012}
F.~Souza, E.~Valle, G.~C{\'a}mara-Ch{\'a}vez, A.~d.~A. Ara{\'u}jo, An
  evaluation on color invariant based local spatiotemporal features for action
  recognition, in: 25th Conference on Graphics, Patterns and Images (SIBGRAPI),
  2012.

\bibitem{Yaghoubyan:2015}
S.~H. Yaghoubyan, M.~A. Maarof, A.~Zainal, M.~F. Rohani, M.~M. Oghaz, Fast and
  effective bag-of-visual-word model to pornographic images recognition using
  the freak descriptor, Journal of Soft Computing and Decision Support Systems
  2~(6) (2015) 27--33.

\bibitem{Zhang:2013}
J.~Zhang, L.~Sui, L.~Zhuo, Z.~Li, Y.~Yang, An approach of bag-of-words based on
  visual attention model for pornographic images recognition in compressed
  domain, Neurocomputing 110~(13) (2013) 145--152.

\bibitem{OpenCV}
G.~Bradski, {The OpenCV Library}, Dr. Dobb's Journal of Software Tools.

\bibitem{Moustafa:2015}
M.~{Moustafa}, {Applying deep learning to classify pornographic images and
  videos}, ArXiv e-prints\href {http://arxiv.org/abs/1511.08899}
  {\path{arXiv:1511.08899}}.

\end{thebibliography}

\end{document}